\documentclass{article}
\usepackage{microtype}
\usepackage{graphicx}
\usepackage{subfigure}
\usepackage{enumitem}

\usepackage{booktabs} 
\usepackage{hyperref}

\usepackage[accepted]{icml2021}

\usepackage{amsthm, amsmath, amsfonts, amssymb}
\usepackage[capitalize,noabbrev]{cleveref}

\newcommand{\E}{\mathbb{E}}
\newcommand{\V}{\mathbb{V}}
\newcommand{\R}{\mathbb{R}}

\newcommand{\indep}{{\perp \!\!\! \perp}}

\graphicspath{{figures/}}

\begin{document}
\twocolumn[
\icmltitle{Return-based Scaling: Yet Another Normalisation Trick for Deep RL}
\begin{icmlauthorlist}
\icmlauthor{Tom Schaul}{dm}
\icmlauthor{Georg Ostrovski}{dm}
\icmlauthor{Iurii Kemaev}{dm}
\icmlauthor{Diana Borsa}{dm}
\end{icmlauthorlist}
\icmlaffiliation{dm}{DeepMind, London, UK}
\icmlcorrespondingauthor{Tom Schaul}{tom@deepmind.com}
\icmlkeywords{Reinforcement learning, adaptive scales, robustness, multiple discounts}
\vskip 0.3in
]
\printAffiliationsAndNotice{}  

\begin{abstract}
Scaling issues are mundane yet irritating for practitioners of reinforcement learning. Error scales vary across domains, tasks, and stages of learning; sometimes by many orders of magnitude. This can be detrimental to learning speed and stability, create interference between learning tasks, and necessitate substantial tuning. We revisit this topic for agents based on temporal-difference learning, sketch out some desiderata and investigate scenarios where simple fixes fall short. The mechanism we propose requires neither tuning, clipping, nor adaptation. We validate its effectiveness and robustness on the suite of Atari games. Our scaling method turns out to be particularly helpful at mitigating interference, when training a shared neural network on multiple targets that differ in reward scale or discounting.
\end{abstract}

\section{Introduction}
\label{sec:intro}

Learning a value function is a central component of most model-free reinforcement learning (RL), represented by neural networks as function approximator of choice in deep RL~\citep{nfq,dqn}.
Training value functions via regression (e.g., a mean squared error loss) is common practice, but unlike in supervised learning there is no standard preprocessing step (such as whitening) that adjusts the \emph{scales} of the learning targets.

\begin{figure*}[tb]
    \centerline{
    \includegraphics[width=0.64\textwidth]{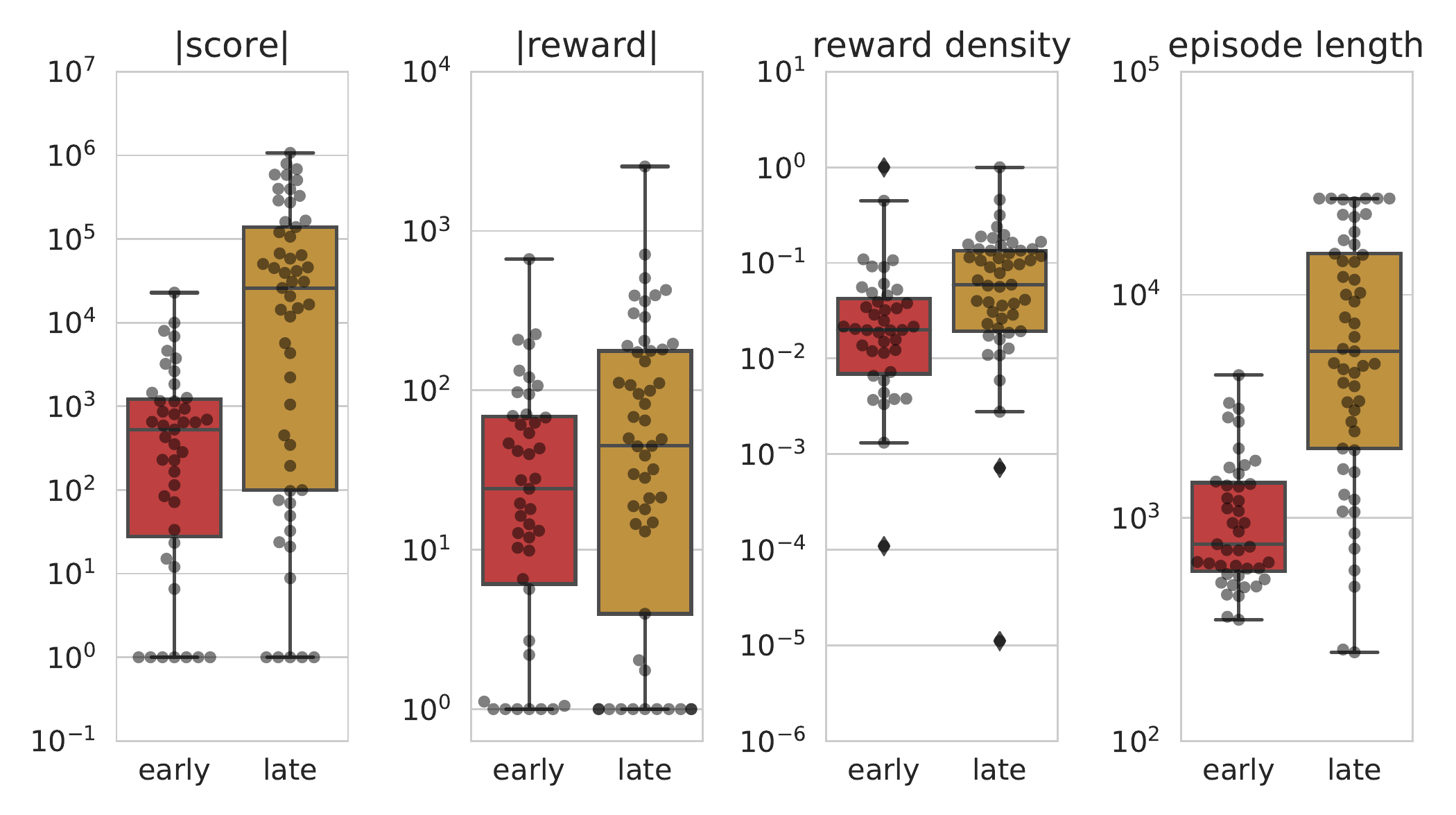}
    \includegraphics[width=0.36\textwidth]{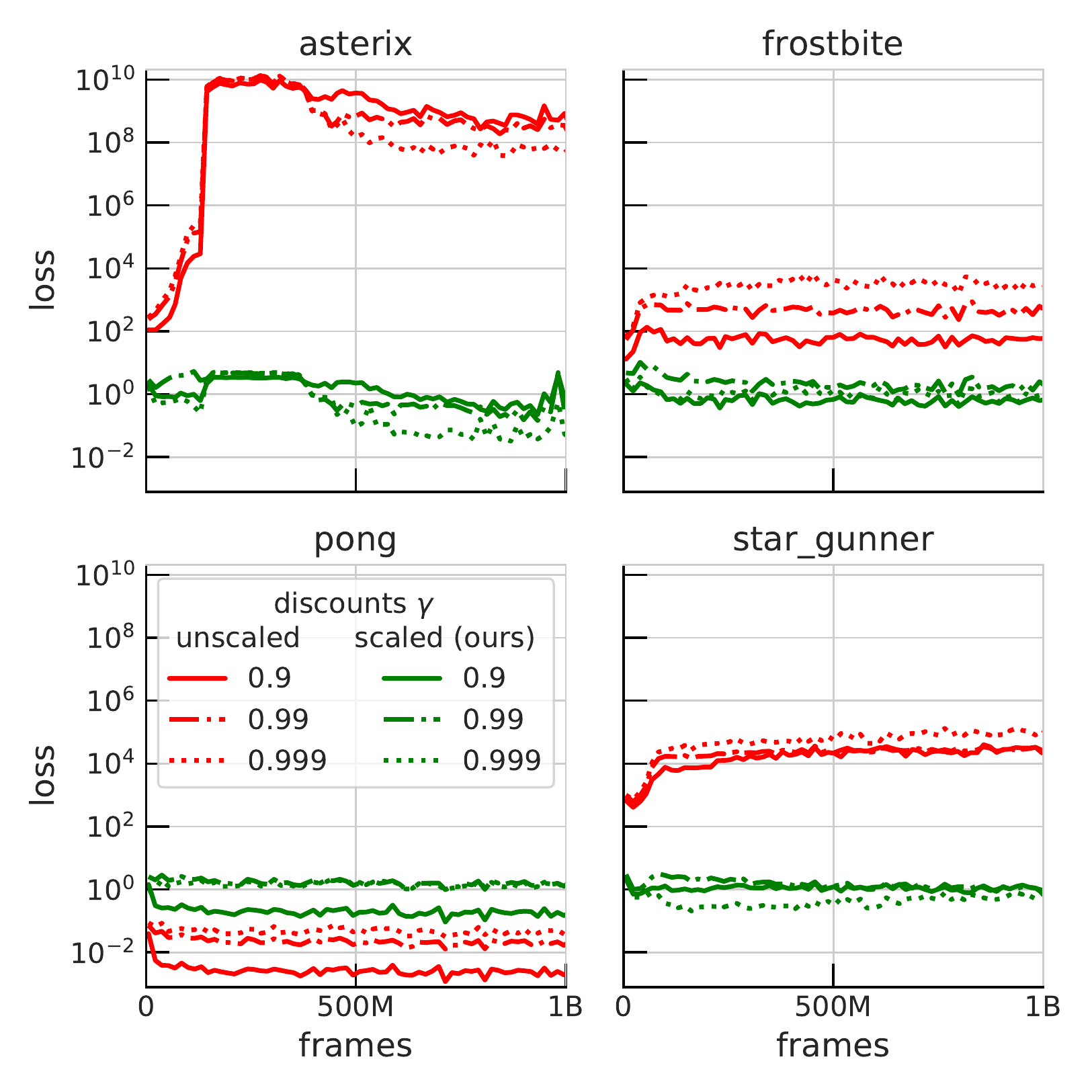}
    }
    \vspace{-1em}
    \caption{
    {\bf Left:} Scale challenges in Atari. Each subplot shows the variability of scales across 57 Atari games (one point per game), for both the early phase of learning (frames $< 10$M, in red) and the late phase ($800$M $<$ frames $< 1$G, in orange). Note how scores (undiscounted returns) vary by many orders of magnitude, which is a compound effect of changing reward scales, reward densities, and episode lengths. Also note how much these statistics can change over the course of learning.
    {\bf Right:}
    Illustration of empirical loss scales on a few individual Atari games. They can vary by 10 orders of magnitude across domains, and can increase or decrease substantially over the course of training (red curves), or both. Each line-style corresponds to a different discount factor $\gamma$. Green curves show how our proposed scaling maps the corresponding scales to a much narrower range. 
    }
    \vspace{-1em}
    \label{fig:challenges}
\end{figure*}

The resulting error scales depend on the reward scales (and density), which can vary widely across tasks or domains, and as values are cumulative quantities, they directly depend on the discount factor too.
On top of this, multiple sources of non-stationarity can cause scales to vary during learning (besides accuracy changes), such as changing policy and data distribution, or discovering new rewards.
In practice, these variations can span many orders of magnitude, see Figure~\ref{fig:challenges} for some realistic examples drawn from Atari.
A further complication arises when training multiple components with differently scaled errors within the same system (e.g., a network with multiple heads), such as an actor and a critic, auxiliary prediction heads, successor features, etc.

\begin{figure}[tb!]
    \centering
    \includegraphics[width=\columnwidth]{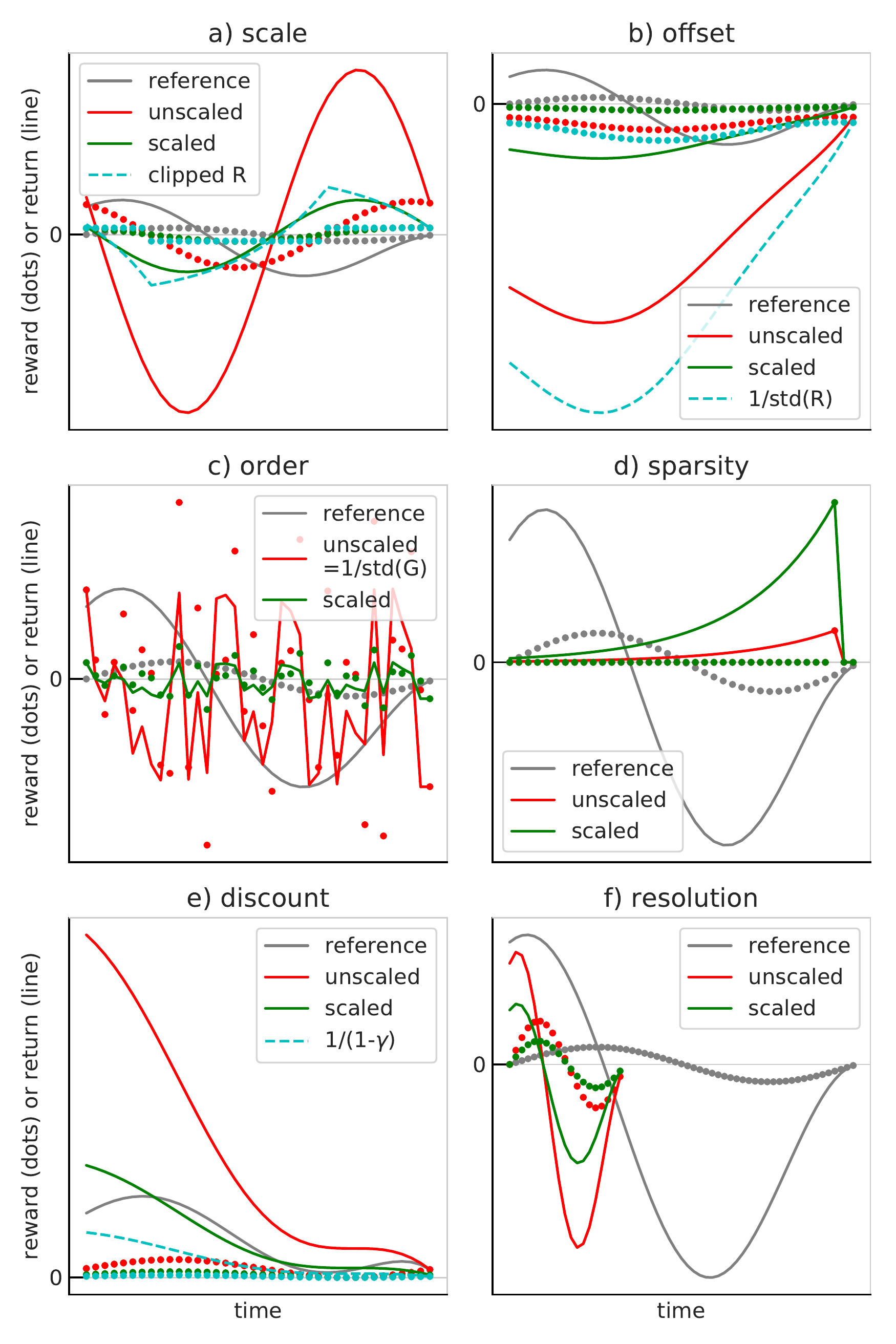}
    \vspace{-2em}
    \caption{Six scenarios for establishing scale intuitions.
    Rewards sequences are shown as dots, return curves as full lines (of matching color). Each panel juxtaposes a reference sequence (in gray) at canonical scale, a second unscaled sequence (in red) and how our method would linearly adjust its scale (in green).
    We encourage the reader to consider whether the match of scales between the green line and the gray line is satisfying, or at least better than the red line.
    In addition, scenarios (a), (b), (c) and (e) highlight a failure mode of some `false friend' (see Section~\ref{sec:false}), shown in cyan. In scenario (c), cyan and red lines overlap exactly, and in scenario (e) red and gray dots overlap.
    \vspace{-1em}
    }
    \label{fig:intuitions}
\end{figure}

This seemingly mundane phenomenon is nevertheless a frequent contributor to thorny practical issues, such as an excessive need for hyper-parameter tuning, instabilities and slowdowns in the learning dynamics, or interference between multiple learning objectives.
It should not come as a surprise, therefore, that mitigation strategies abound in the field of deep RL; some of these do not take center stage, but come in the guise of `tricks of the trade'. Examples include reward and gradient clipping~\citep{dqn,impala}, discount factors, non-linear reward or value transforms~\citep{dqfd,pohlen2018observe,non-linear-transforms}, separations between value and advantage~\citep{dueling}, or separate networks instead of a shared torso~\citep{agent57}. Methods that tackle the issue of scale head-on include~\citep{popart,dann2021adapting}, see Section~\ref{sec:discussion} for more.

In this work, we revisit the topic of scale, lay out some prototypical scenarios, spell out criteria for desirable normalisation schemes, and show where simple fixes fall short (Section~\ref{sec:intuitions}). 
Based on those insights we then propose a preprocessing-like scaling mechanism of our own, dubbed \emph{return-based scaling}, 
which is algorithm-agnostic as it requires no access to agent internals.
It sidesteps undesirable side-effects, has low computational and implementation complexity, and does not introduce any new hyper-parameters (Section~\ref{sec:proposal}). 
A suite of experiments on Atari validates its effectiveness and robustness, and shows a substantial performance gain when used to balance the scales of multiple learning objectives (Section~\ref{sec:experiments}).

\section{Intuitions}
\label{sec:intuitions}

In this section, we discuss six toy examples to illustrate that a robust scaling method needs to produce sensible results in quite diverse scenarios. 
We reduce the question to what \emph{linear} factor would be appropriate for a canonical scale. The objects of interest are sequences of rewards $R_t$ ($t\in\{1,\ldots,T\}$), and the sequences of \emph{returns} $G_t$ that are obtained when combining rewards with a discount factor $\gamma$:
\[
G_t := R_t + \gamma G_{t+1} = \sum_{t'=t}^T \gamma^{t'-t} R_{t'},
\]
where $T$ is the length of the episode.
The desired product of learning is the \emph{value} $V_t:=v(S_t):=\E[G_t|S_t]$, i.e., the expected return from a state $S_t$, which in deep RL is approximated by a neural network.

We consider several scenarios, varying a different aspect of a reference reward sequence in each, 
illustrate how that affects the scales of rewards and returns, and show how our 
proposed method would rescale things in order to put the modified sequence on the same footing as the reference.

\begin{itemize}[topsep=4pt,itemsep=0pt]
    \item The purest scenario is when all rewards (and thus returns) are linearly scaled up or down by some factor: we expect a reasonable scaling mechanism to correct by the same factor (Figure~\ref{fig:intuitions}a).
    
    \item A first subtlety arises when rewards are \emph{offset} additively instead of multiplicatively: in that case, their variance is identical, but the scale of returns can vary significantly (see Figure~\ref{fig:intuitions}b). 
    Reward offsets introduce an additional value-learning burden, as they require estimating the time until episode termination; and depending on offset magnitude, this aspect can dominate.
    
    \item The scales of returns on their own are not a sufficient characteristic of the desirable scale, the sequential structure matters as well. This is illustrated in Figure~\ref{fig:intuitions}c, where the comparison sequence has identical returns to the (smooth) reference, but in shuffled order (adjusting rewards to produce such a sequence): this preserves the variance of returns but can dramatically change error scales, potentially adding large and sudden jumps in value on top of a (presumably) smooth state sequence, which can increase the difficulty of learning in a smooth function approximator like a neural network~\citep{nva}.
    
    \item Similarly, reward scales on their own are insufficient. For example, Figure~\ref{fig:intuitions}d contrasts a dense reward and a sparse reward situation (with the same maximal reward), and how our scaling method boosts the scale factor to compensate for sparsity.
    
    \item Another dimension to take into account is the discount $\gamma$, which influences how reward scales relate to return scales, as well as the number of future rewards to be considered (none for $\gamma=0$, all for $\gamma=1$). Figure~\ref{fig:intuitions}e indicates that the same reward sequence, accumulated under different discounts, may need different scaling.
    
    \item Finally, changing the time-resolution of the sequence (while preserving total reward) is not a neutral operation and will generally require rescaling, see Figure~\ref{fig:intuitions}f. This can show up when using `action repeats'~\citep{dqn} or options~\citep{options}.
\end{itemize}

\subsection{False friends}
\label{sec:false}
These toy scenarios intimate that a simple one-component scaling mechanism may well fall short. 
But it may be instructive to make things more concrete, by illustrating the failure modes of some minimalist ideas that quickly come to mind: these often look tempting when considering one of the above scenarios in isolation, but fail in others.
\begin{itemize}[topsep=4pt,itemsep=0pt]
    \item Normalising rewards, by dividing them by their standard deviation $\operatorname{std}(R)$: by construction, this ignores both reward offsets (2b) and discounts (2e).
    \item Normalising returns, by dividing them by their standard deviation $\operatorname{std}(G)$: this ignores sequential structure and treats smooth sequences the same as erratic ones (2c).
    \item Normalising time-scales, by down-scaling returns by the effective time-horizon $1/(1-\gamma)$: this can be overly conservative, because while the maximal error grows with the horizon, the \emph{typical} error is often much smaller\footnote{We revisit this point in more depth in the appendix.} (2e).
    \item Clipping rewards, while commonly used to address scale issues, is problematic because it does not preserve the semantics of the original problem (2a), and can make it impossible to attain optimality.
\end{itemize}

\subsection{Point of scaling}
So far, we have not committed to the specific quantity to be rescaled. In fact there are numerous viable choices on where to apply a scaling factor: we could rescale rewards at the source, bootstrap targets, errors, losses, gradients, or parameter updates -- each of which comes with advantages and disadvantages.
In a simple agent (e.g., that optimises a squared error with vanilla SGD) many of these are equivalent. In practice however, agents may employ additional components such as Huber losses, gradient clipping, adaptive optimisers~\citep{adam}, off-policy corrections, or prioritised experience replay~\citep{per}. In order for a scaling method to be as agent-agnostic as possible, it is thus preferable to apply the scale factor as far upstream as possible.
On the other hand, it may be important to preserve the true semantics of the prediction targets, namely produce values at their original scales\footnote{For example, because values are subsequently combined, such as via successor features~\citep{sf}, or drive a softmax policy with a specific temperature. Note also that this desideratum of preserved value semantics may not always apply, e.g., when all that matters is finding the action with the highest value.}.
This leads us to prefer a point of scaling that is furthest upstream yet semantics-preserving, namely rescaling \emph{errors}.

\section{Proposal: Return-based scaling}			
\label{sec:proposal}

Our aim is utterly pragmatic. Any method we devise must satisfactorily address the scale issues of the scenarios above, while being as simple as possible: in particular, we only consider simple linear rescaling methods that do not have any hyper-parameters (so that the tuning effort is not just shifted) and can be quickly implemented in a wide range of agents.
Under these constraints, our objective cannot be to obtain ideal learning dynamics or optimisation properties -- rather our method is better seen as a data preprocessing step that is complementary to adaptive optimisers such as Adam~\citep{adam}, or to within-network normalisations such as Batch-norm or Layer-norm~\citep{batchnorm,layernorm}.
	
Our starting point is the temporal-difference (TD) error $\delta$, which takes the form
\[
\delta_t = R_t + \gamma_t V'_{t+1} - V_t,
\]
the simplest case of a one-step transition from $S_t$ to $S_{t+1}$.
The target value $V'$ can be different from $V$, for example when using a target network with different parameters, or when constructing it from Q-values, as in Q-learning: $V'_{t+1} = \max_a Q(S_{t+1}, a)$.

We propose to replace raw TD-errors $\delta_t$ by a scaled version 
\begin{equation}
    \bar{\delta}_t := \frac{\delta_t}{\sigma},
    \label{eq:scaled}
\end{equation}
where $\sigma\in\R^+$ is an adaptive scale factor.
The next subsection derives an approximation of $\sigma$ based on only reward and return statistics.

\subsection{Derivation of an approximate scale}
\label{sec:derivation}

For determining overall error scales, the regime of interest is the \emph{transient} regime, long before convergence: this is what happens in early learning, as well as when learning in the presence of changing policies and constantly new data.
Whereas errors eventually approach zero (given sufficient capacity and a convergent data distribution), the errors in the transient regime are the ones that characterise the \emph{problem}.

We can write the variance of TD-errors in the following way
\begin{eqnarray}
\V[\delta] 
\nonumber&=& \V[R + \gamma V' - V]
\\\label{eq:V1}&=& \V[R + \gamma (V' - V) - (1-\gamma) V]
\end{eqnarray}
where $\V[X]$ denotes the variance of a random variable $X$, and time indices are omitted.
We assume the (approximate) independence relations 
$R \indep \gamma (V'-V) \indep (1-\gamma) V$, noting that the first  is justified only in early learning (i.e., the transient regime), and the second  is based on value \emph{gaps} being generally uninformative about value \emph{magnitudes}. Equation~\ref{eq:V1} then decomposes into
\begin{eqnarray}
\V[\delta] &\approx& \V[R] + \V[\gamma (V' - V)] + \V[(1-\gamma) V].
\label{eq:Vdec1}
\end{eqnarray}
The variance of a product of independent variables obeys
\begin{eqnarray*}
\V[XY]&=& \E[X]^2\V[Y]+\V[X]\E[Y]^2 + \V[X]\V[Y] 
\\&=& \E[X]^2\V[Y]+\V[X](\E[Y]^2 + \V[Y]) 
\\&=& \E[X]^2\V[Y]+\V[X]\E[Y^2],
\end{eqnarray*}
so
when assuming\footnote{Note that, perhaps unconventionally, we treat $\gamma$ as a random variable here, because it is zero at the final step of an episode (even when it is constant throughout the episode).}
\footnote{Of course a higher discount leads to larger values, but for a given discount function, there is not much correlation between $V_t$ and (per-transition) $\gamma_t$, except that both are jointly zero on the terminal transition.}
$\gamma \indep V'$ and $\gamma \indep V$ (and using $\V[1-\gamma]=\V[\gamma]$) we obtain
\begin{eqnarray}
\V[\delta] &\approx& \V[R] + \bar{\gamma}^2 \V[V' - V] + \V[\gamma] \E[(V' - V)^2]
\nonumber\\&&
+ (1- \bar{\gamma})^2 \V[V] + \V[\gamma] \E[V^2],
\label{eq:Vdec2}
\end{eqnarray}
where we denote $\bar{\gamma}:= \E[\gamma]$.
It is reasonable to assume that values take on similar overall scales to returns, very early in learning. So to a first approximation, return statistics can take the place of value statistics, i.e., $\V[V] \approx \V[G]$ and $\E[V^2] \approx \E[G^2]$, thus
\begin{eqnarray}
\V[\delta] &\approx& \V[R] + \bar{\gamma}^2 \V[G' - G] + \V[\gamma] \E[(G' - G)^2]
\nonumber\\&&
+ (1- \bar{\gamma})^2 \V[G] + \V[\gamma] \E[G^2]
\label{eq:VtoG}
\end{eqnarray}
One way to approximate the statistics of one-step differences
$G'-G = R-(1-\gamma)G$ is to use analogous (approximate) independence assumptions as above:
\begin{eqnarray*}
\E[(G' - G)^2]
&=&  \E[(R-(1-\gamma)G)^2]
\nonumber
\\&\approx&  \E[R^2] + (1-\bar{\gamma})^2 \E[G^2] 
\\&&
- 2  (1-\bar{\gamma}) \E[R] \E[G]
\\&\approx&  \E[R^2] + (1-\bar{\gamma})^2 \E[G^2] 
\\&&
- (1-\bar{\gamma})^2 \E[G]^2
- \E[R]^2
\\&=& \V[R] + (1-\bar{\gamma})^2 \V[G] 
\label{eq:e2vgap}
\end{eqnarray*}
where the last approximation uses\footnote{This is motivated by $G$ being the $\gamma$-discounted sum of $R$: if $\gamma$ and $R$ were constant, then $G = R/(1-\gamma)$ exactly. When $R$ is a random variable that has a relatively homogeneous structure in time (e.g., not all rewards concentrated at one end of the sequence), which we assume to be the case, then the approximation is valid.} 
$\E[R] \approx (1-\bar{\gamma}) \E[G]$.
Similarly
\begin{eqnarray*}
\V[G'-G] 
&\approx& 
\V[R] + (1-\bar{\gamma})^2 \V\left[G\right]+ \V[\gamma] \E\left[G^2\right].
\label{eq:varvgap}
\end{eqnarray*}
Substituting these into Equation~\ref{eq:VtoG} gives
\begin{eqnarray}
\V[\delta] 
\nonumber&\approx& \V[R] 
+ \bar{\gamma}^2\V[R] 
+ \bar{\gamma}^2(1-\bar{\gamma})^2 \V[G]
\\&&\nonumber
+ \bar{\gamma}^2\V[\gamma] \E[G^2]
+ \V[\gamma] \V[R] 
\\&&\nonumber
+ \V[\gamma](1-\bar{\gamma})^2 \V[G] 
\\&&\nonumber
+ (1-\bar{\gamma})^2 \V[G] + \V[\gamma] \E[G^2]
\\&=&\nonumber
(1+\bar{\gamma}^2+\V[\gamma] ) \V[R]
\\&&\nonumber
+(1+\bar{\gamma}^2+\V[\gamma]) (1-\bar{\gamma})^2 \V[G]
\\&&\nonumber
+(1+\bar{\gamma}^2)\V[\gamma]\E[G^2]
\\&\approx&\nonumber
\V[R] + (1-\bar{\gamma})^2 \V[G]
+\V[\gamma]\E[G^2]
\\&\approx&
\V[R] +\V[\gamma] \E[G^2] 
\label{eq:Vfinal}
\end{eqnarray}
where the last approximation exploits the fact that the $\V[R]$ term dominates the 
$(1-\bar{\gamma})^2\V[G]$ term\footnote{We also dropped the $1+\bar{\gamma}^2+\V[\gamma] \in [1,3]$ and $1+\bar{\gamma}^2 \in [1,2]$ factors, which are small constant factors that matter only marginally when scales span multiple orders of magnitude.}.
This final expression is simple to estimate in practice and we will use it as our scale factor for `return-based scaling':
\begin{equation*}
    \sigma^2 := \V[R] +\V[\gamma] \E[G^2] \approx \V[\delta].
\end{equation*}
It turns out that this is sufficient to satisfactorily address all of the scenarios discussed in Section~\ref{sec:intuitions}, because its three components have sufficient information about reward scales, discounting, offset, etc. In fact, the green curves in Figure~\ref{fig:intuitions} were rescaled using exactly Equation~\ref{eq:Vfinal}.

\begin{figure}[tb]
    \centering
    \vspace{-0.5em}
    \includegraphics[width=\columnwidth]{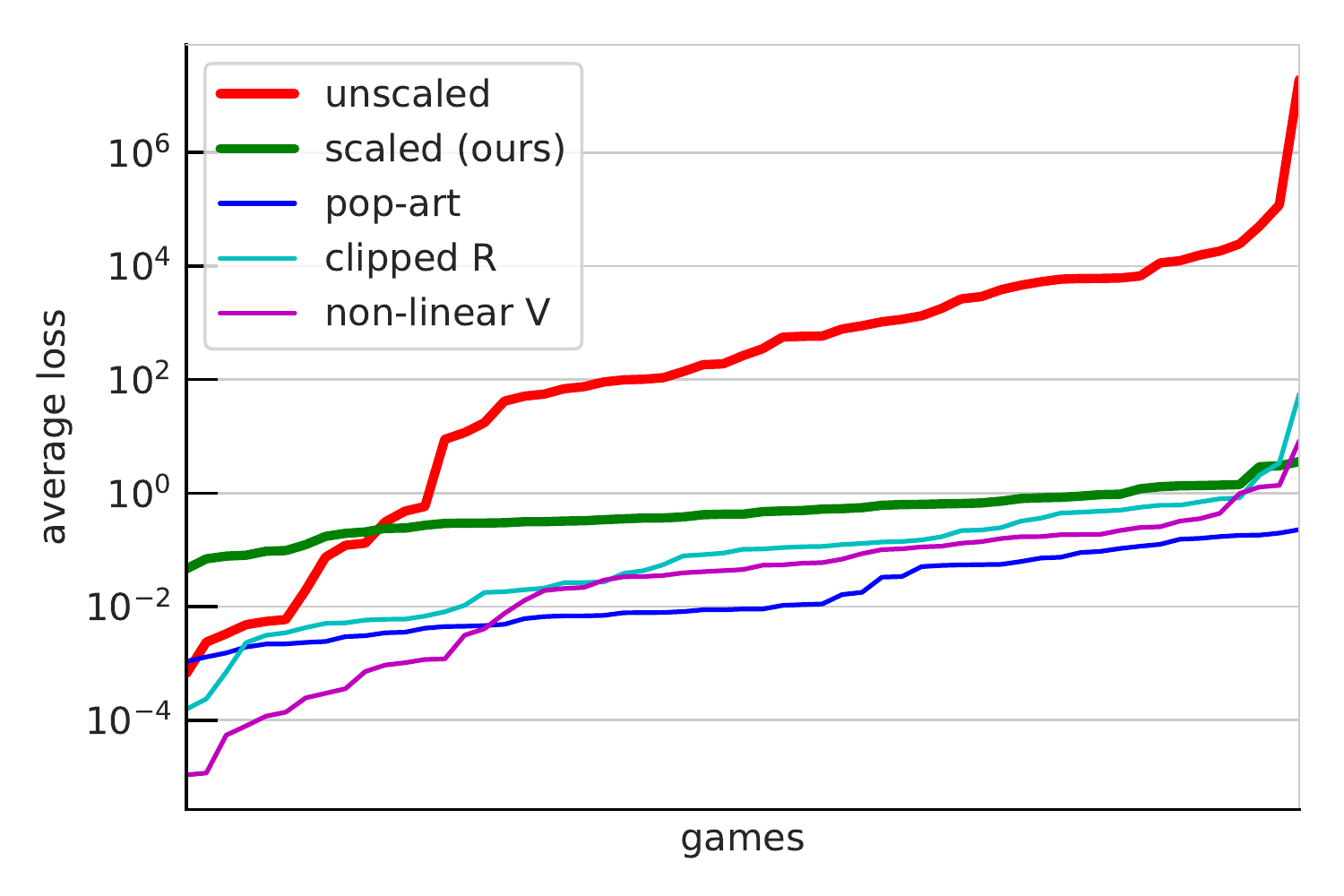}
    \vspace{-2.8em}
    \caption{Loss scales across 57 Atari games, when using return-based scaling (green) or not (red). Data is averaged across training, and the 57 per-game averages are sorted before plotting, i.e., a handful of games have (unscaled) average losses below $10^{-2}$, as well as a handful above $10^{4}$. Note how return-based scaling brings all loss scales into a narrow band. Thin lines show the corresponding results for three commonly used alternative scaling methods, each of which spans a wider range.
    \vspace{-1.8em}
    }
    \label{fig:all-scales}
\end{figure}

\subsection{Implementation}		
In keeping with our aim of not adding any hyper-parameters, we propose to estimate the statistics $\V[R]$, $\V[\gamma]$ and $\E[G^2]$ based on all data the agent has ever seen. This is a conservative approach that we prefer for robustness and simplicity, even though conceivably a faster time-scale of \emph{tracking} scale statistics in non-stationary environments could lead to further gains. We leave such investigations to future work.

Another practical design choice arises around initialisation, before sufficient data has been seen to make $\sigma$ stable and reliable. 
Concretely, two edge cases need to be addressed. First, the case where no reward has \emph{ever} been seen, in this case Equation~\ref{eq:scaled} would lead to a division by zero: one option to prevent this is to use 
$\bar{\delta}_t := \delta_t/\max(\sigma,\sigma_V)$ where $\sigma_V$ is the noise level on the value function at neural network initialisation (typically $\sigma_V \approx 10^{-2}$).

Second, the moment a reward arrives that is much larger than all previously encountered ones (e.g., the first non-zero reward): it is imperative that this enters the scale statistics \emph{before} a learning update based on it takes place (otherwise the update magnitude could be enormous). This is not difficult to guarantee in a synchronous, single-stream RL agent, but (speaking from experience) deserves some care in distributed, replay-based or asynchronous settings.
Specifically, we use $\bar{\delta}_t := \delta_t/\max(\sigma,\sigma_V,\sigma_{\text{batch}})$, where $\sigma_{\text{batch}}$ is applying Equation~\ref{eq:Vfinal} on just the transitions in the current batch of data -- this term underestimates scales in general (because replayed return sequences are generally truncated before the end of episode), and is therefore most often ignored anyway, but it resolves the stability issue for this edge case.

\section{Atari experiments}
\label{sec:experiments}

\begin{figure*}[tb]
    \centerline{
    \includegraphics[width=0.95\textwidth]{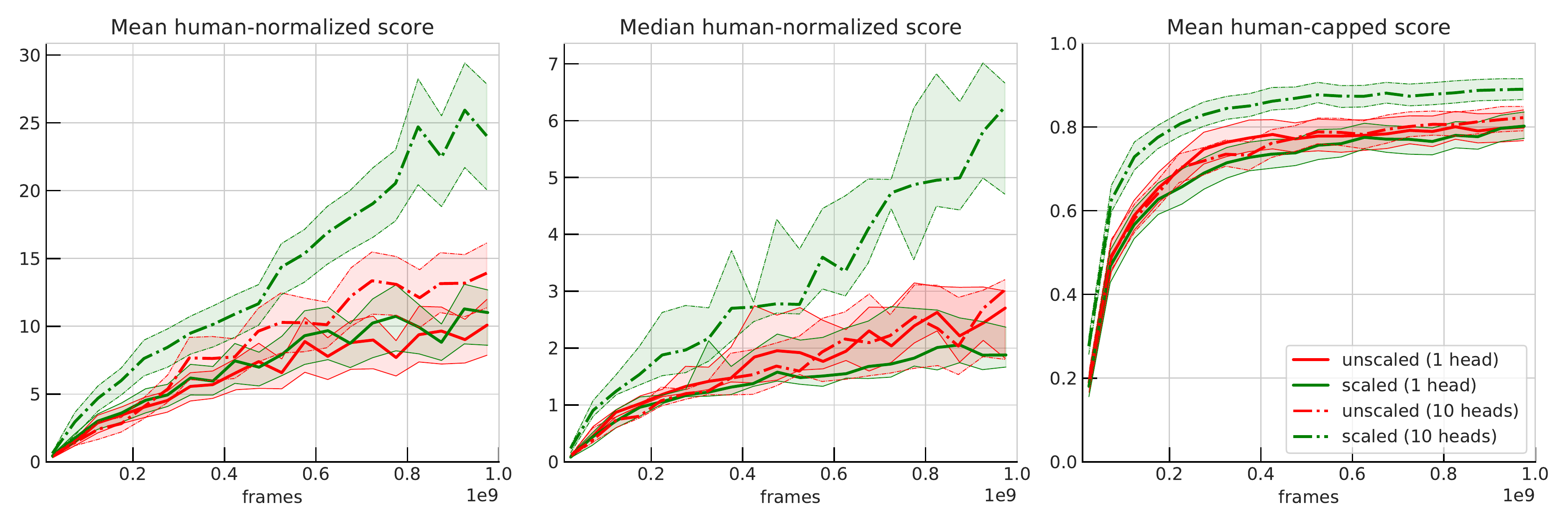}
    }
    \vspace{-1em}
    \caption{Aggregate performance results across 57 Atari games (1 seed), see Figure~\ref{fig:atari-all} (appendix) for per-game details. The four variants show unscaled (red) and scaled (green) results, for both the single head (solid lines) and 10-head (dash-dotted lines) scenarios. Shaded areas indicate inter-quartile ranges computed via bootstrap sampling the games (indicating sensitivity to scores in individual games). We find that return-based scaling has a massive benefit for the 10-head setup, for any metric considered, but is on par with the unscaled baseline in the 1-head setup. Also, the 10-head setup  demonstrates its benefit over the single-head one only when the losses of the different heads are appropriately balanced via return-based scaling, but collapses to essentially 1-head performance otherwise.}
    \vspace{-1em}
    \label{fig:atari-summary}
\end{figure*}

\begin{figure}[tbh]
    \centering
    \includegraphics[width=\columnwidth]{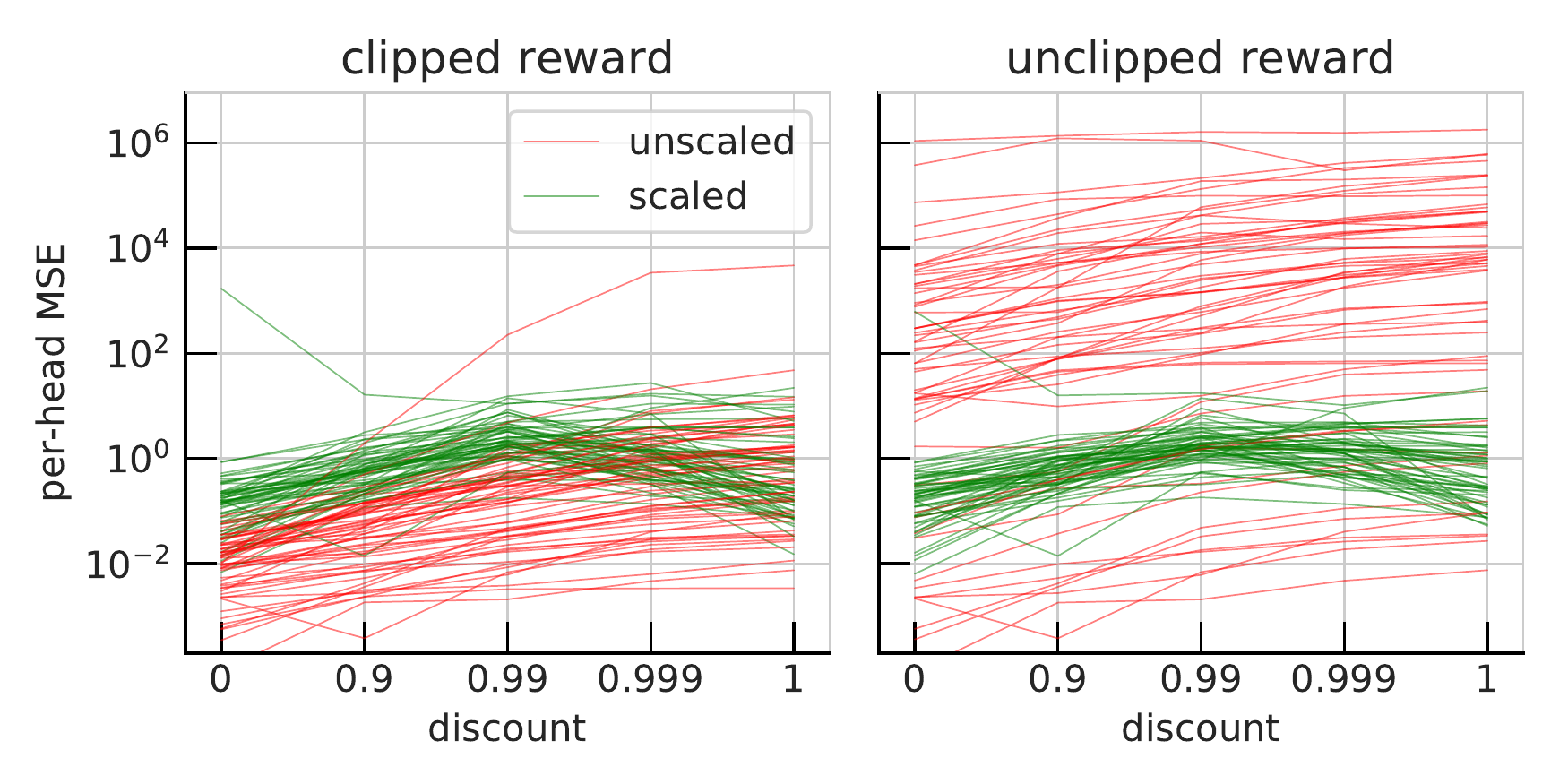}
    \vspace{-2em}
    \caption{Average per-head loss scales, when using return-based scaling (green) or not (red). Each line corresponds to one of the 57 Atari games, connecting the per-head loss scales of two groups (clipped and unclipped rewards) of 5 heads with different discounts.
    Note how the unscaled losses vary significantly across discounts (typically about 100x), and the difference in scales between clipped and unclipped rewards is even bigger.
    \vspace{-1.5em}
    }
    \label{fig:discount-scales}
\end{figure}

The toy results in Section~\ref{sec:intuitions} are a good sanity check, but as an eminently practical method, we need to validate it in \emph{realistic} deep RL contexts where scales matter.
The suite of 57 Atari games~\cite{ale} is highly appropriate; on the one hand it is widely studied and comes with strong baseline agents, and on the other hand it has vast diversity of scales. In fact, as Figure~\ref{fig:challenges} shows, its rewards, reward densities, episode lengths and resulting total scores vary across many orders of magnitude; what is more, all of these change significantly across training. On top of this challenging domain, we consider value accumulation horizons all the way from $\gamma=0$ (one-step) to $\gamma=1$ (undiscounted).

\subsection{Setup}

As agent architecture, we opt for an implementation of R2D2~\citep{r2d2}, which is representative of high-performing distributed value-based RL algorithms\footnote{But without the full complexity of its descendants like Agent57~\citep{agent57}.}. 
Its overall baseline performance (unscaled) is roughly on par with published results, but note that we operate in a much smaller data regime (1B frames instead of 30B), so `final' scores are not comparable.
Appendix~\ref{app:setup} gives details of the experimental setup and all hyper-parameters.

We investigate two setups, in each case comparing return-based scaling to an unscaled baseline. The vanilla (1 head) case is an R2D2 agent with a single discount that optimises for discounted unclipped reward (with unchanged default $\gamma=0.997$). The multi-head variant is different in two ways: its neural network has 10 separate heads that estimate value functions for 5 different discounts $\gamma \in \{0, 0.9, 0.99, 0.999, 1\}$, 5 heads for unclipped reward, and 5 heads for clipped reward\footnote{By \textit{clipped} reward we refer to constraining the reward to lie in $[-1, 1]$ as proposed first in \citep{dqn}, whereas \textit{unclipped} reward refers to the raw reward as provided by the environment.}. All experience collected is used to train all heads with multi-step Q-learning (with no further off-policy correction). The second difference is how the heads are used to generate experience: each episode, a bandit picks the head most likely to generate high (undiscounted, unclipped) returns (as in~\citep{adx-bandit-arxiv}). The same per-head performance statistics also determine which head's policy is executed for evaluation.
When return-based scaling is applied, it is done separately for each head.

\subsection{Results}
\label{sec:results}

The first order of business is to validate that return-based scaling does indeed produce similar error and loss scales in all these scenarios. Figure~\ref{fig:challenges}(right) hinted at this already, and Figure~\ref{fig:all-scales} demonstrates it across the board, for 57 games, 5 discounts, and all stages of learning.
On its own, this is a valuable result, because it implies that scale-related tuning and stability questions become obsolete, and one confounding factor has been eliminated.

Next, we establish that the impact of our method on overall performance is either neutral (if the baseline setup handled scales in a satisfactory way) or beneficial (if it did not). 
The top-line results are shown in Figure~\ref{fig:atari-summary}, which shows aggregate performance (human-normalised mean, median, and capped mean) across all 57 Atari games, trained on $10^9$ frames. Interestingly, the outcome is very different for the two setups. 
Perhaps surprisingly, in the 1-head case, return-based scaling does not lead to a meaningful performance difference. The main reason for this is likely that as a high-performing agent on Atari, the unscaled baseline must have been designed and tuned to handle the scale differences sufficiently well. In fact, the main mechanism at work here is the adaptive optimiser Adam~\citep{adam}: we look at this in more depth in Appendix~\ref{app:adam}.

\begin{figure*}[tb]
    \centerline{
    \includegraphics[width=0.9\textwidth]{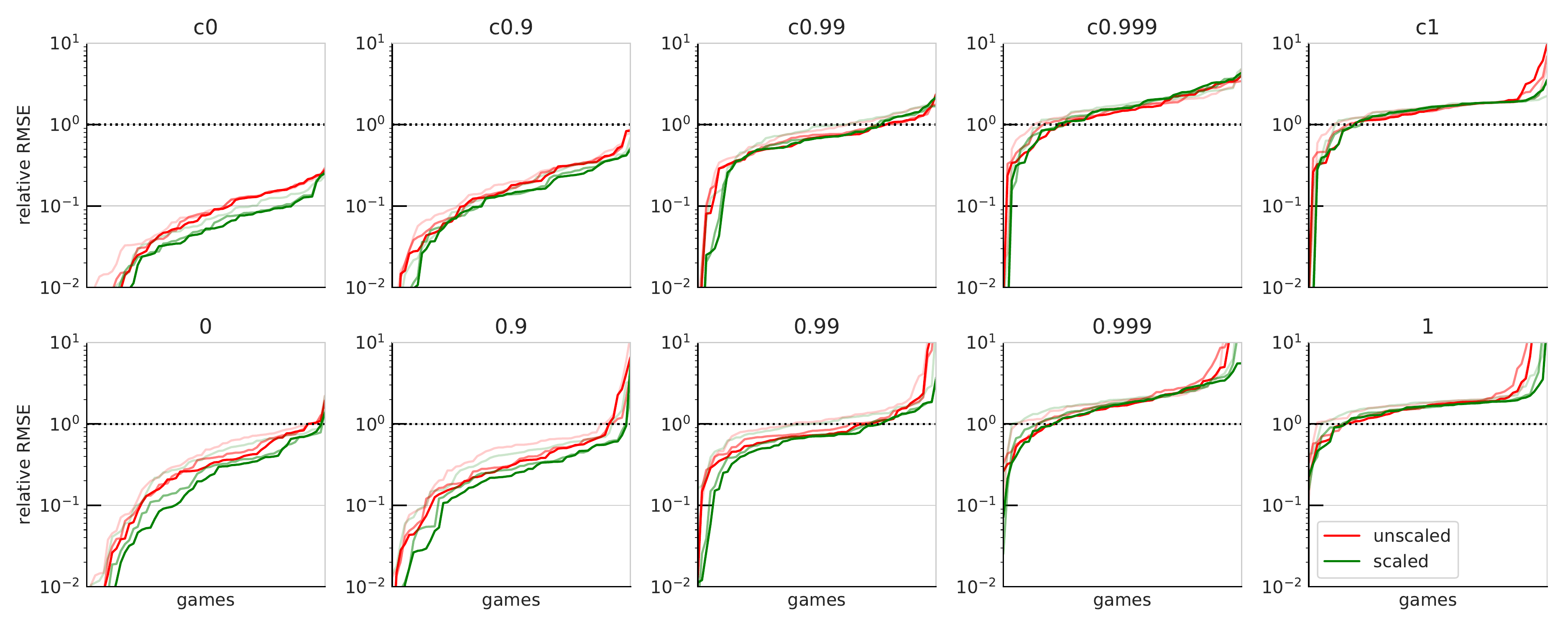}
    \vspace{-1.5em}
    }
    \caption{Summary of value accuracies across games (sorted horizontally). Relative root mean squared errors (RMSE) are normalised by $\V[G]$, i.e., the best constant value function should have a \emph{relative} RMSE of $1$ (this should cancel out most effects due to underlying policies being different). Solid lines show final accuracy, faint lines show accuracies at earlier stages of training. In aggregate, return-based scaling (green) leads to better value accuracy in all heads compared to the baseline (red), with the largest gaps visible on low-discount heads.
    We hypothesise that return-based scaling's balanced loss contributions across heads is responsible for this benefit (see also Figure~\ref{fig:err-loss-frac}).
    \vspace{-1em}
    }
    \label{fig:imp-accuracy}
\end{figure*}

For the 10-head case, return-based scaling leads to a large performance increase compared to the unscaled 10-head baseline (as well as all 1-head experiments). This indicates that better \emph{relative} scales of the many heads are beneficial to learning: Figure~\ref{fig:discount-scales} shows that error scales are indeed well-balanced across discounts, and nearly identical whether learning about clipped or unclipped rewards.
Figure~\ref{fig:imp-accuracy} validates that a better balance of relative loss scales leads to better value accuracy for \emph{all} heads.
Additionally (in the appendix), Figure~\ref{fig:err-loss-frac} shows the relation between a head's value accuracy and its contribution to the overall loss. 

\subsection{Comparisons to other scaling methods}
\label{sec:compare}

\begin{figure}[tb]
    \centering
    \includegraphics[width=\columnwidth]{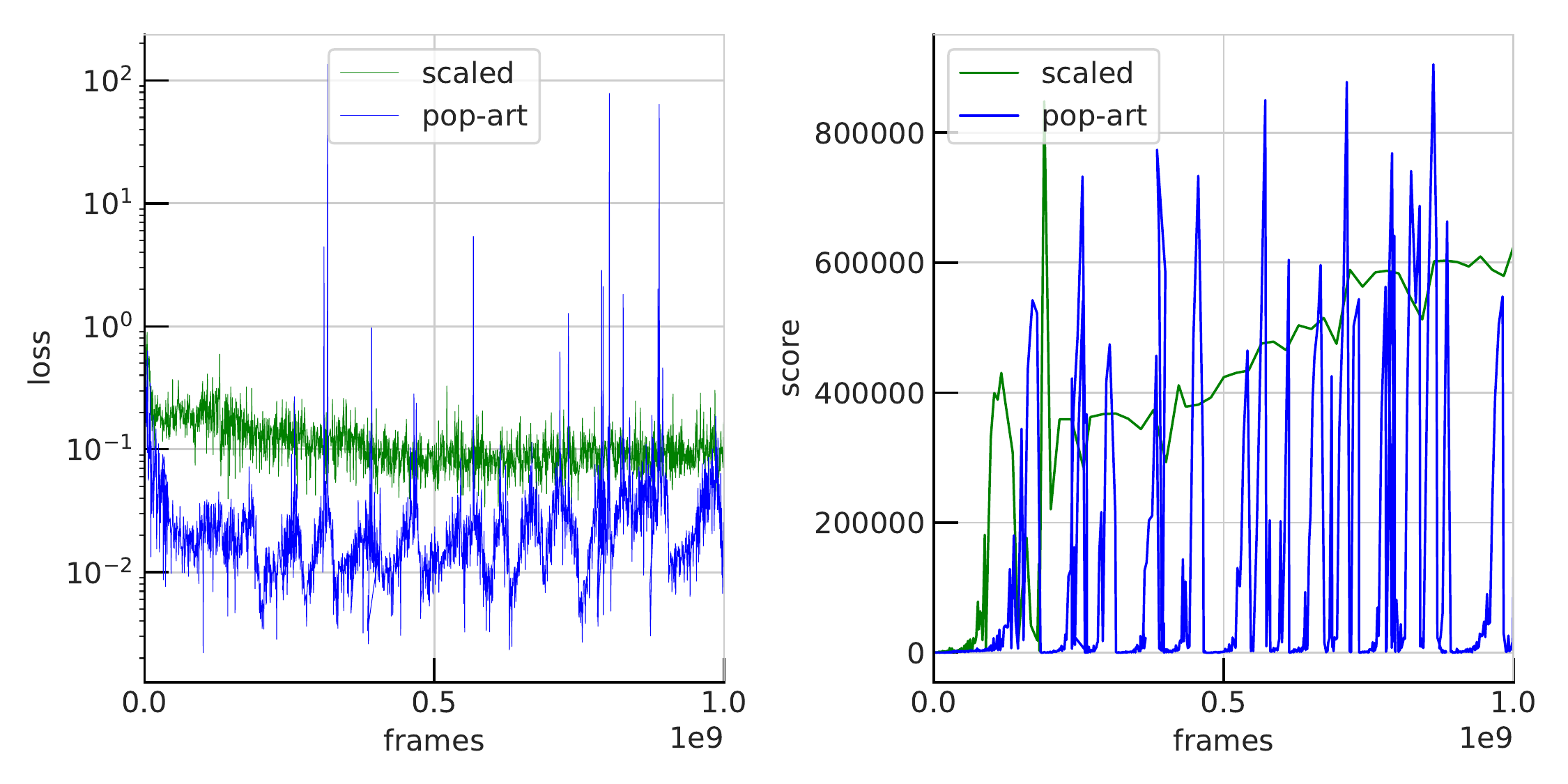}
    \vspace{-2em}
    \caption{Stability in the presence of large reward spikes. The left plot shows the loss scales across learning (subsampled, not smoothed) for our method and Pop-Art on the game of \textsc{Asterix} (10-head setup). The right plot shows the corresponding evaluation scores. This shows that our method produces very stable loss scales throughout, despite massive variations in score, resulting in stable learning. In contrast, Pop-Art is adapting scales on a much more rapid time-scale, leading to large jumps, which in turn result in repeated performance collapses.
    \vspace{-1em}
    }
    \label{fig:popart-asterix}
\end{figure}

We also conducted some head-to-head comparisons with alternative scaling methods from the literature:
\begin{itemize}[topsep=4pt,itemsep=0pt]
    \item Reward clipping, i.e.~capping rewards to lie in $[-1, 1]$, is commonly used in Atari since at least DQN~\citep{dqn}. It breaks the original problem semantics, as the agent becomes blind to large reward events, making some games impossible to solve (e.g., \textsc{Bowling} or \textsc{Skiing}) or imposing a performance ceiling. However, the heuristic of accumulating many rewards independently of their magnitude happens to be well-aligned with the design of many Atari games, so overall results tend to be good\footnote{Note that our 10-head setup exploits this heuristic as well by permitting the agent to learn about and to pursue clipped rewards, whenever that is advantageous \emph{in terms of unclipped} total score. Looking at Figures~\ref{fig:atari-summary} and~\ref{fig:1head}, this seems to explain a good chunk of the 10-head performance improvement.}.
    Figure~\ref{fig:clip-unclip} (appendix) shows how the trade-off between pursuing clipped and unclipped rewards plays out in some example games.
    
    \item Pop-Art~\citep{popart, popart2} is similar in spirit to our approach, but it differs in a number of ways: it normalises the bootstrap targets (not the errors), based on recent (not all-time) statistics. It also has multiple tunable hyper-parameters, and is more invasive implementation-wise, as it requires write-access to the neural network's last layer weights to do a secondary update after the gradient descent step. Its property of tracking scales over a short horizon can also lead to instability\footnote{\textsc{Asterix} has a peculiar reward structure: the score counter is limited to 6 digits, and resets when it hits 1'000'000. To the agent this is a one-step reward of -1'000'000. Such an outlier event is rare, and disruptive to any agent. The ideal policy is to stop obtaining rewards (just) before the score resets.}, as shown in Figure~\ref{fig:popart-asterix}.
    
    \item Non-linear value transforms~\citep{non-linear-transforms} affect scales in a non-linear way: they squash large-magnitude errors while not affecting small-magnitude ones, and shift the emphasis of learning.
\end{itemize}
Figure~\ref{fig:all-scales} summarises the scaling effect of these three comparable methods (1-head setup): they all mitigate the range of loss scales, but not as much as return-based scaling.
In terms of performance, the main difference is that reward clipping leads to a significant increase in mean score over the others, see Figure~\ref{fig:1head} (appendix).
For the 10-head setup, we also compared our method to Pop-Art (see Figure~\ref{fig:popart}, appendix), which does almost as well as return-based scaling, even if it lags far behind in terms of median score.
Reward clipping is not a directly comparable method, as half of the 10 heads already have clipped rewards.

\begin{figure}[tb]
    \centerline{
    \includegraphics[width=0.5\columnwidth]{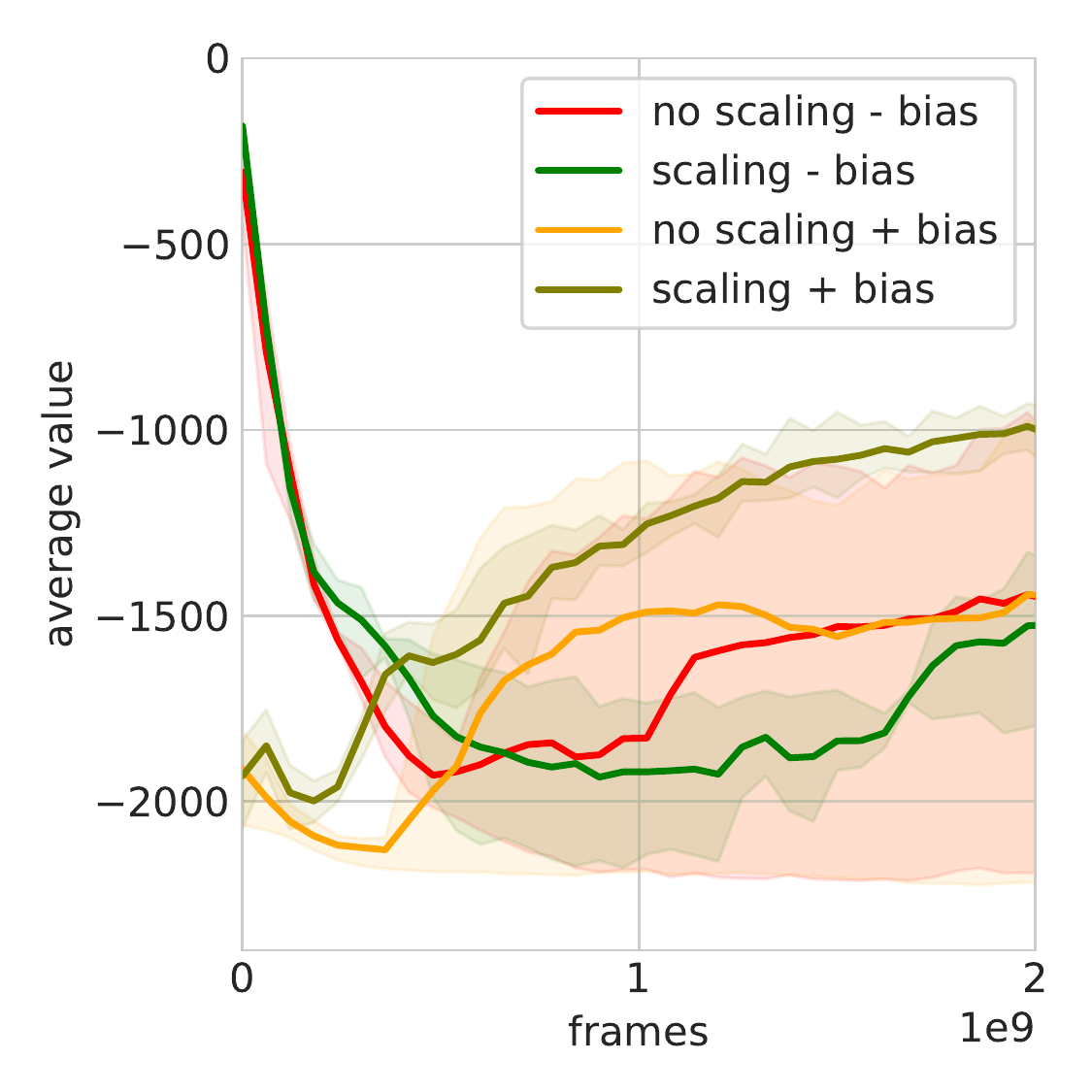}
    \includegraphics[width=0.5\columnwidth]{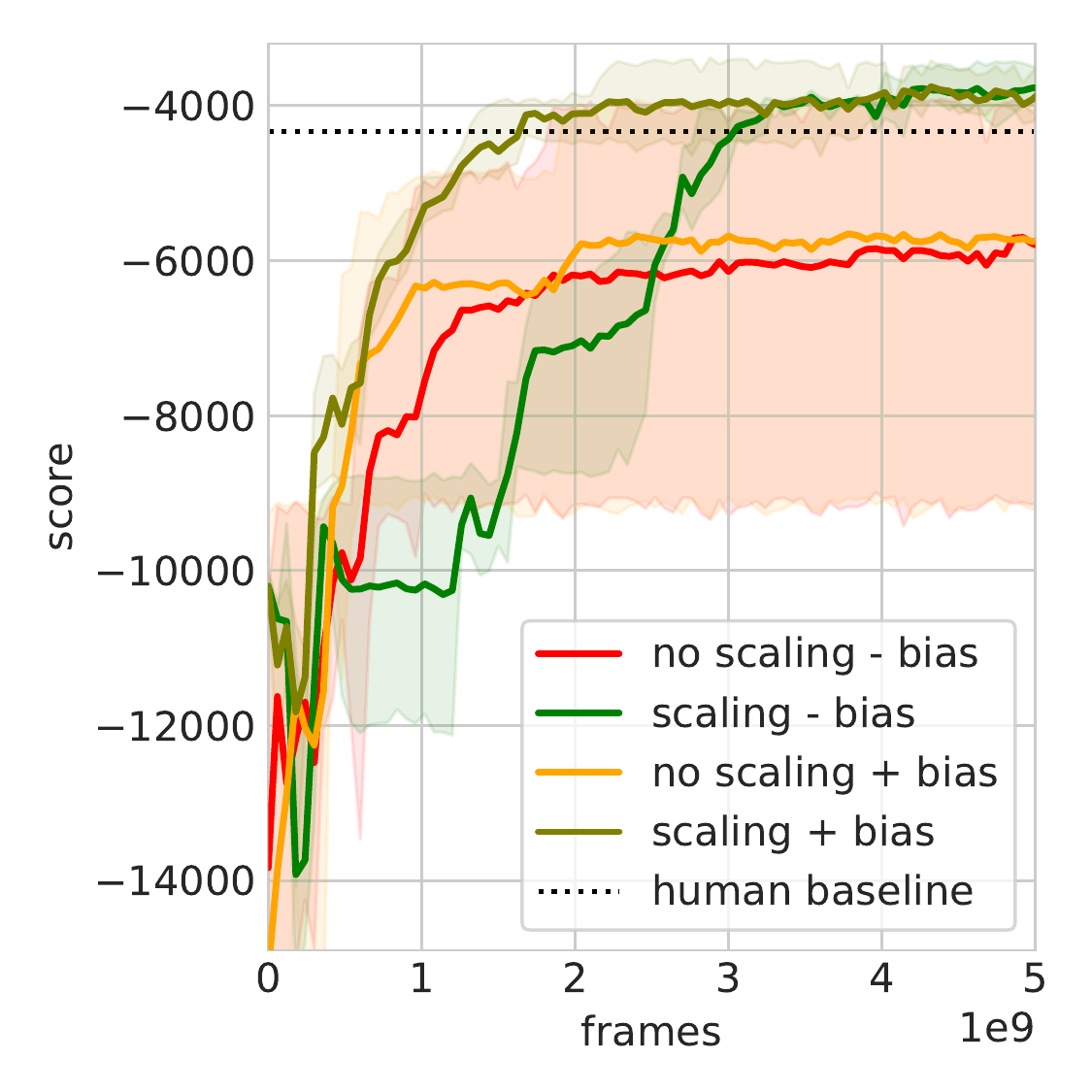}
    }
    \vspace{-1em}
    \caption{Initializing value bias, a case-study on the game \textsc{Skiing} (3 seeds per setting, shaded areas show min-max span). This game has an unusual reward scheme, with large negative rewards each step. This makes initialising the value near zero (as is standard for deep learning) problematic: as the left subplot shows, it can take 500M frames until the average value is accurate. Contrast this with the analogous experiments that differ only in how the bias weight of the value function is initialised (in orange and olive-green, denoted ``+ bias''). Without this reward-offset related delay in value learning, overall performance takes off faster too, surpassing human-level after 2B frames; this compares favourably with e.g. Agent57~\citep{agent57}, which required 78B.
    \vspace{-1em}
    }
    \label{fig:skiing}
\end{figure}

\section{Discussion}
\label{sec:discussion}

\paragraph{Return-based versus error-based scaling.}
When considering the starting point of our derivation (Equation~\ref{eq:V1}), an obvious alternative approach comes to mind, namely to simply measure the actual TD-errors encountered during training and use their variance for normalisation. 
There are a few pragmatic arguments for using return-based statistics: as \emph{problem-side} quantities they do not depend on agent-specific internals, they are not affected by learner competence, value initialisation, or optimisation dynamics.
More importantly, however, return-based scaling does not suffer from the risk of \emph{noise amplification}, a phenomenon where as the value accuracy keeps improving, error-based rescaling continually amplifies whatever residual approximation error is left, possibly hitting numerical instabilities (see Figure~\ref{fig:noise-amplif}).
The noise amplification is also a nuisance in a multi-head setup, where the up-scaling of the noise in simple-to-learn heads (e.g., $\gamma=0$) will interfere with the learning of the others.
Error-based scaling is similar to Pop-Art, and as we have seen in Figure~\ref{fig:popart-asterix}, this can lead to unstable dynamics.

\paragraph{Interplay with regularisation.}
Another potentially\footnote{We have not investigated this yet, because our baseline agent has no form of regularisation loss.} important aspect is whether a scaled TD-error interacts differently with regularisation terms (such as $L2$ \citep{farebrother2018generalization} or entropy regularisation) than an unscaled one. We hypothesise tuning trade-offs between the value loss and other loss components will become easier -- especially if value losses change massively over time, in which case any fixed coefficient on other loss components would normally make them either dominant or irrelevant.

\paragraph{Initial value offset.}
When rewards (and thus returns) are significantly offset from zero, it may be beneficial to initialise the bias weight of the value network to an appropriate value, namely $\E[G]$, based on some initial statistics (this is a light-weight implementation change when return statistics are collected anyway). We would expect this to make little difference in most Atari games, unless reward offsets are substantial, such as in the game of \textsc{Skiing}. Preliminary experiments (see Figure~\ref{fig:skiing}) indicate that this value bias initialisation trick can indeed speed up learning, resulting in human-level performance on \textsc{Skiing} in under $2$B frames.

\paragraph{Target networks.}
When using a target network, the gap $V'-V$ can be substantially larger than expected because $V'$ may lag behind $V$. It is possible to take this into account in the derivation in Section~\ref{sec:derivation} and to obtain the following expression instead:
$    \sigma^2 := \V[R] +\V[\gamma] (\E[G^2] + \Delta V^2)$,
where $\Delta V^2 := \E[(V'(S_t)-V(S_t))^2]$ is the mean squared difference between online and target networks, computed on the same states.
Empirically, this term tends to be dominated by the $\E[G^2]$ term, so we omit it from our proposed method (keeping it simpler and more agent-agnostic).

\paragraph{Correlations across multiple heads.}
So far, we have treated each objective in the multi-head setup as completely separate. While simple, this approach may be suboptimal when the information is correlated or redundant. For future work, we expect there could be significant further gains when normalising across all heads jointly, akin to \emph{whitening}. 

\paragraph{Trading off multiple losses.}
Balancing the learning of many optimisation problems that share a common representation is a common scenario in both supervised learning and RL. It can occur in a multitask setting, when we are interested in multiple predictions \citep{gvf,sf}, or as auxiliary losses to shape a shared representation \citep{jaderberg2016reinforcement}. The latter has been shown to benefit the primary task, due to the implicit transfer, and it is commonly used in RL to compensate for the sparsity of reward signal \citep{stooke2020decoupling}. Moreover in RL, some work have used auxiliary tasks for better exploration \citep{riedmiller2018learning,colas2019curious,ngu}. Despite these benefits, the effectively training a system with multiple losses requires a lot of care, especially when loss scales differ from each other or vary across time.
A large body of work has investigated different ways of adapting the individual losses' contributions based on various learner-based quantities: the gradient norm \citep{chen2018gradnorm}, compatibility across gradients \citep{lin2019adaptive}, task uncertainty \citep{kendall2018multi}, or average past losses \citep{hu2019learning}. Similarly, methods like population-based training\citep{pbt} and meta-gradients \citep{xu2018meta,zahavy2020selftuning} can be employed to adaptively adjust the coefficients between losses over time. Although very general, these tend to be quite expensive. Finally, it is important to note that all these methods, and others that look at the learning dynamics and various learner-side quantities, can have complementary benefits, and could be use in conjunction with the return-based scaling proposed here.

\section{Conclusion}
We have introduced \emph{return-based scaling}, a method that uses a linear scale factor based purely on problem-side statistics (rewards, discounts, returns)
to rescale TD-errors into an optimization-friendly range.
We discuss its appropriateness to most common scenarios as well as edge cases, 
and demonstrate its effectiveness in a practical and realistic setting (R2D2 on Atari), both in terms of scaling effects and overall performance.
The most noteworthy result is how beneficial the rescaling is when learning multiple values of different scale within the same system: our method balances in their contributions in a way that leads to large overall performance improvements.
As it adds no tunable hyper-parameters and is simple to implement, we expect it to be applicable out-of-the-box to various other value-based RL algorithms.

\subsection*{Acknowledgements}
The authors are grateful for the insights and feedback provided by Arthur Guez, David Silver, John Quan, Simon Osindero, David Szepesvari, David Amos, Hado van Hasselt, Claudia Clopath, and Miruna P\^{i}slar and the wider DeepMind team.

\bibliography{bib}
\bibliographystyle{icml2021}

\clearpage
\appendix
\section{Detailed experimental setup}
\label{app:setup}

\subsection{Agent}

The agent used in our Atari experiments is a distributed implementation of a value- and replay-based RL algorithm 
derived from the Recurrent Replay Distributed DQN (R2D2) architecture \citep{r2d2}. 
This system comprises of a fleet of $192$ CPU-based actors 
concurrently generating experience and feeding it to a distributed experience replay buffer, 
and a single GPU-based learner randomly sampling batches of experience sequences from replay
and performing updates of the recurrent value function by gradient descent on a suitable RL loss. 

The value function is represented by a convolutional torso feeding into a linear layer, followed 
by a recurrent LSTM \citep{hochreiter1997long} core, whose output is processed by a further linear layer before finally being 
output via a Dueling value head \citep{dueling}. The exact parameterization follows the 
slightly modified R2D2 presented in \citep{ez-greedy}, see Table \ref{tab:hyper} for a full
list of hyper-parameters.
It is trained via stochastic gradient descent on a multi-step TD loss 
(more precisely, a $5$-step Q-learning loss) with the use of a periodically updated
target network \citep{dqn} for bootstrap target computation, using minibatches of sampled replay sequences.
Replay sampling is performed using prioritized experience replay \citep{per} 
with priorities computed from sequences' TD errors following the scheme introduced in \citep{r2d2}. 
As in R2D2, sequences of $80$ observations are used for replay, 
with a prefix of $20$ observations used for burn-in. In a slight deviation from the original, our agent
uses a fixed replay ratio of $1$, i.e. the learner or actors get throttled dynamically if the average number
of times a sample gets replayed exceeds or falls below this value; this makes experiments more reproducible and stable.

Each actor periodically pulls the most recent network parameters from the learner to be used
in its $\varepsilon$-greedy policy. Instead of assigning a fixed value of $\varepsilon$ to each actor, 
the actors randomly sample values of $\varepsilon$ at the beginning of each episode, with a distribution
mimicking that used in \citep{r2d2}.
In addition to feeding the replay buffer, all actors periodically report their reward, discount and return
histories to the learner, which then calculates running estimates\footnote{In our experiments it was important for these estimates to have double precision (\textsc{float64}) to prevent a loss of precision when computing variances.} of $\V[R]$, $\V[\gamma]$ and $\E[G^2]$
to perform return-based scaling.

Differently from most past Atari RL agents following DQN \citep{dqn}, our agent uses the raw 
$210 \times 160$ RGB frames as input to its value function (one at a time, without frame stacking),
though it still applies a max-pool operation over the most recent 2 frames to mitigate flickering
inherent to the Atari simulator. As in most past work, an action-repeat of $4$ is applied,  
episodes begin with a random number of no-op actions (up to $30$) being applied, and 
time-out after $108$K frames (i.e. $30$ minutes of real-time game play).

\begin{table*}[t]
    \centering
    \begin{tabular}{c|c}
        
        \textbf{Neural Network} \\
         Convolutional torso channels & $32, 64, 128, 128$ \\
         Convolutional torso kernel sizes & $7, 5, 5, 3$ \\
         Convolutional torso strides & $4, 2, 2, 1$ \\
         Pre-LSTM linear layer units & $512$ \\
         LSTM hidden units & $512$ \\
         Post-LSTM linear layer units & $256$ \\
         Dueling value head units & $2 \times 256$ (separate linear layer for each of value and advantage) \\

         \hline
        \textbf{Acting} \\
         Number of actors & $192$  \\
         Action repeats & $4$ \\
         Actor parameter update interval & $400$ environment steps \\
         $\varepsilon$ distribution (for $\varepsilon$-greedy policy) & 
           Uniform $\sim \{0.0005, 0.0015, 0.005, 0.015, 0.05, 0.15, 0.4\}$ \\
         
         \hline
         \textbf{Replay} \\
         Replay sequence length & $80$ (+ prefix of $20$ in burn-in experiments) \\ 
         Replay buffer size & $4\times 10^6$ observations ($10^5$ part-overlapping sequences) \\
         Priority exponent & $0.9$ \\
         Importance sampling exponent & $0.6$ \\ 
         Fixed replay ratio & $1$ update per sample (on average) \\
         
         \hline
         \textbf{Learning} \\
         Discount $\gamma$ & $0.997$ (for 1-head setup)\\
         Mini-batch size & $32$ \\
        
         Optimizer \& settings & Adam \citep{adam}, \\
         & learning rate $\eta = 2\times10^{-4}$, $\epsilon=10^{-6}$, \\
         & momentum $\beta_1=0.9$, second moment $\beta_2=0.999$ \\
         Target network update interval & $400$ updates \\
         Gradient clipping & not used \\
         Huber loss threshold & not used \\

    \end{tabular}
    \caption{Atari agent hyper-parameter values.}
    \label{tab:hyper}
\end{table*}

Our agent is implemented with JAX \citep{jax2018github}, uses the Haiku \citep{haiku2020github}, Optax 
\citep{rlax2020github}, Chex \citep{chex2020github}, and RLax \citep{optax2020github} libraries for neural networks, optimisation, testing, and RL losses, 
respectively, and Reverb \citep{Reverb} for distributed experience replay.

\subsection{Training \& evaluation protocol}

All our experiments ran for $200$K learner updates. With a replay ratio of $1$, sequence length of $80$
(adjacent sequences overlapping by $40$ observations), a batch size of $32$, and an action-repeat of $4$ this corresponds 
to a training budget of $200000 \times 32 \times 40 \times 1 \times 4 \approx 10^9$ environment frames
(which is  $\sim 30$ times fewer than the original R2D2). In wall-clock-time, one such experiment takes about $12$ hours.

For evaluation, a separate actor (not feeding the replay buffer) is running alongside the agent using a 
greedy policy ($\varepsilon = 0$), and pulling the most recent parameters at the beginning of each episode.
We follow standard evaluation methodology for Atari, reporting mean and median `human-normalised' scores as
introduced in \citep{dqn} (i.e. the episode returns are normalised so that $0$ corresponds 
to the score of a uniformly random policy while $1$ corresponds to human performance),
as well as the mean `human-capped' score which caps the per-game performance at human level. 
All experiments are conducted across $57$ games, using one seed per game, unless stated otherwise.

\subsection{Hyper-parameter tuning}
We were fortunate to start this investigation from a well-tuned baseline agent code-base, and our method itself has no tunable hyper-parameters. Nevertheless, in early iterations (on a small subset of games), we looked at a few scale-related hyper-parameters, such as learning rate, Adam-$\epsilon$, and prioritisation exponents: around the reported default values (Table~\ref{tab:hyper}) they are not very sensitive. Adam-$\epsilon$ is the one hyper-parameter where different settings reported in prior work (around $10^{-3}$) did not work well in our setup, neither in 1-head nor in 10-head settings, and neither with nor without return-based scaling.

\subsection{Multi-head experiments}

In experiments using multiple value heads, the neural network architecture is shared between the different
value heads up to (and including) the post-LSTM linear layer, with each Dueling value head consisting of its
own advantage and value streams (each with a single hidden layer of $256$ units). Similarly to \citep{dueling},
gradients flowing from each of the $n$ individual value heads to the shared layers are scaled by a constant $1/\sqrt{n}$.

The $10$ value heads in our experiments correspond to the $10$ value functions resulting from using $2$ different reward
functions (clipped and unclipped rewards) and $5$ different discount rates ($\gamma \in \{0, 0.9, 0.99, 0.999, 1\}$).
The value heads are all trained simultaneously from the same sampled experience batches. 

The choice of value head for use in the $\varepsilon$-greedy policy of each actor
is renewed at each episode beginning and follows the method introduced in \citep{adx-bandit-arxiv}:
the agent is supplied with an additional unit which receives all experience generated by all actors, 
executes a bandit algorithm to select value heads to be used in future episodes,
and communicates those selections back to the actors. Following \citep{adx-bandit-arxiv}, 
the bandit's fitness function operates based on the undiscounted returns of the unclipped reward function.
For evaluation, the best value head with respect to the same criterion is used greedily.

The value statistics are accumulated separately for each value head, and determines the scaling factor for that head.

\subsection{Pop-Art experiments}

For our Pop-Art comparisons, we endow our agent's value function with a Pop-Art normalisation following \citep{popart}.
The hyper-parameters involved were not tuned, but fixed to sensible default values:
the statistics step size is fixed at $0.001$, and the scale parameter is bounded below and above, for stability, 
by $0.001$ and $1000$, respectively. In multi-head experiments, the Pop-Art normalisation statistics are
accumulated and applied separately for each of the value heads.

\subsection{Non-linear value transforms}
Experiments for this comparison are modifying the baseline agent in a single point, namely by transforming the bootstrap targets using the `signed hyperbolic' squashing function 
\[
f(x) = \operatorname{sign}(x) (\sqrt{|x|+1}-1),
\]
see~\citep{non-linear-transforms}.

\section{Scale invariance due to Adam}
\label{app:adam}
\begin{figure}[tbh]
    \centering
    \includegraphics[width=\columnwidth]{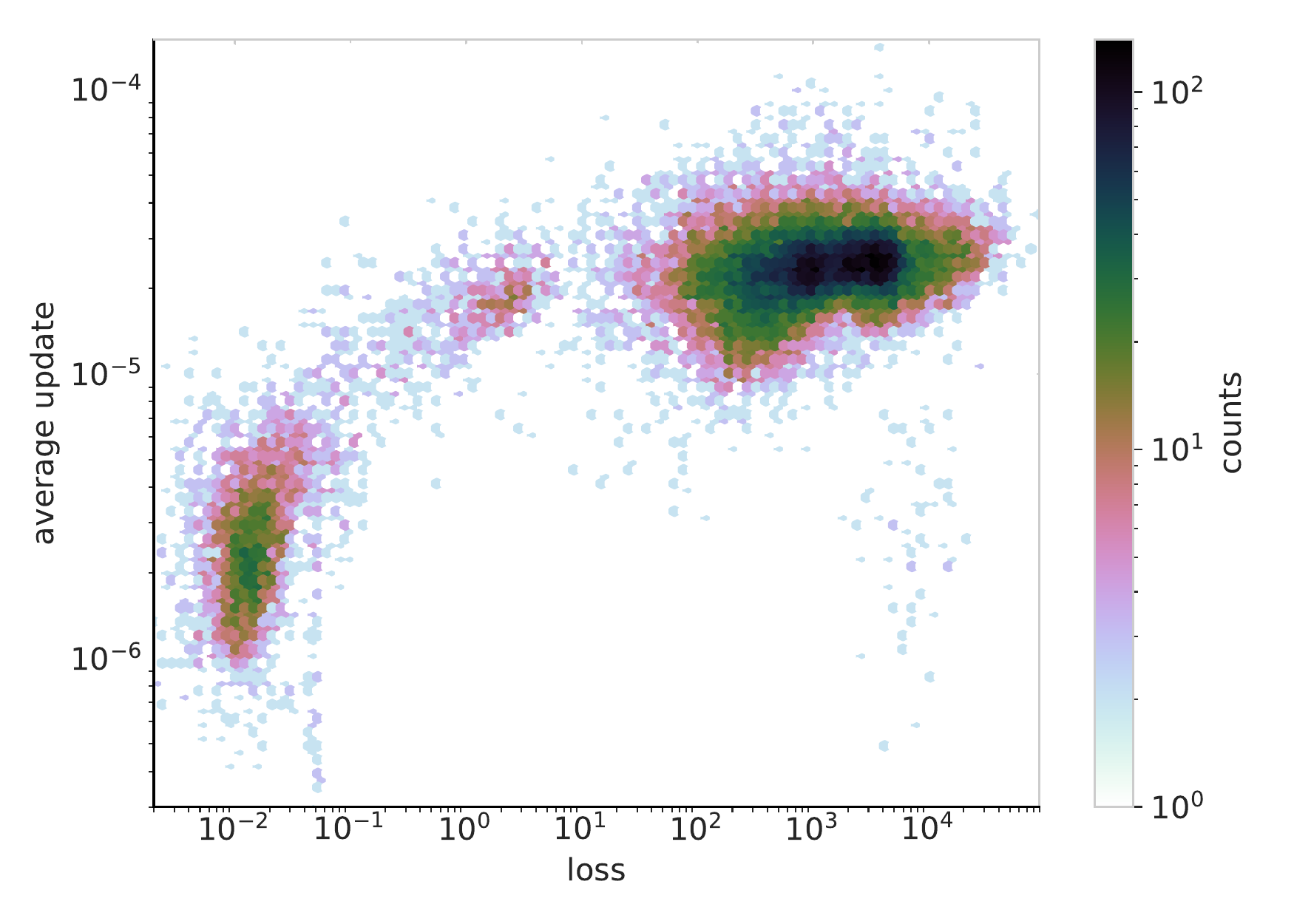}
    \vspace{-3em}
    \caption{Relation between loss scales and parameter update scales after passing through the Adam optimizer (with learning rate $\eta=0.0002$ and $\epsilon=0.001$). Data is collected across a subset of 12 Atari games. Adam normalises the overall update scale, but there is asymmetry: small losses (below 1 or so) lead to smaller updates, while all larger ones are scaled to the same approximate scale. The epsilon parameter is what determines this switching point.
    \vspace{-1em}
    }
    \label{fig:adam}
\end{figure}

As discussed in Section~\ref{sec:results}, a lot of the scale differences in the 1-head setup seem to be neutralised by Adam~\citep{adam}, the adaptive optimisation algorithm employed.
In fact, Adam normalises the update of each parameter by an estimate of its gradient's (recent) standard deviation, an idea related to our proposal, even if applied on the parameter level instead of the error level.
Figure~\ref{fig:adam} illustrates the effect of this in practice. While the effect is somewhat complex, to a first approximation Adam does seem to rescale updates to within a small range, at least when losses are large.

We have experimented in the space of simpler, non-rescaling optimisers like SGD (in an otherwise similar agent setup), however the overall performance is abysmal (with performance hovering around random, even when sweeping over learning rates), and so it is difficult to argue that such an agent constitutes a valid baseline for studying normalisation.

\begin{figure}[tb]
    \centering
    \includegraphics[width=\columnwidth]{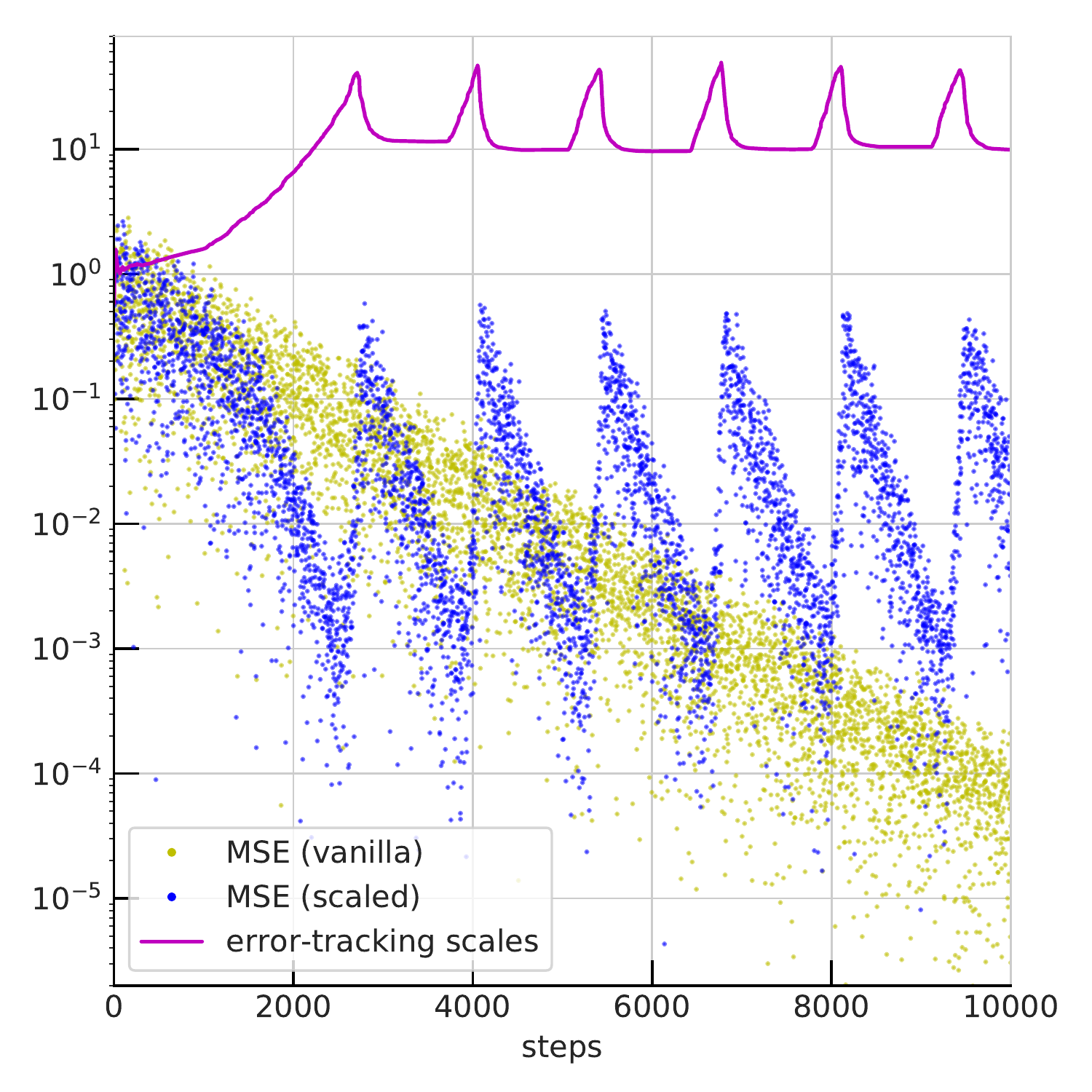}
    \vspace{-2em}
    \caption{Illustrating the noise amplification effect of an error-tracking scaling mechanism.
    The setting here is simple linear regression (MSE loss, updated with SGD, step-size of $10^{-3}$) toward a constant target of zero, with 100-dimensional Gaussian noise as input; in other words the ideal learned weights are all exactly zero. The vanilla version (yellow dots) does exactly this. 
    A simple error rescaling method that is based on recent error statistics (blue dots, scale factors in purple) has different behaviour:
    initially, it speeds up learning, but then it hits an instability, where the (modest) amplification via the scale factor leads to a 1000x jump in error. This pattern then keeps repeating itself forever, and the weights never converge.
    Return-based scaling is not shown, but it would look exactly like vanilla (because all `rewards' and `returns' are zero, it would recover a constant $\max(\sigma, \sigma_{V}) = \sigma_{V} \approx 1$).
    \vspace{-1em}
    }
    \label{fig:noise-amplif}
\end{figure}

\begin{figure*}[t]
    \centering
    \includegraphics[width=\textwidth]{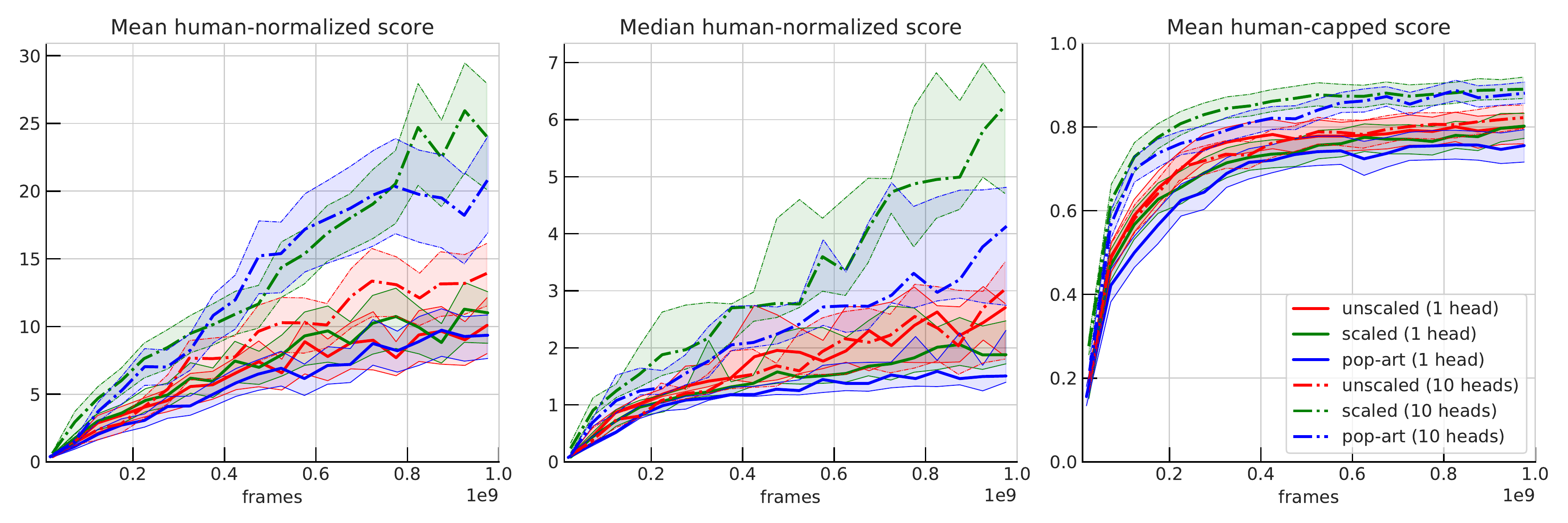}
    \vspace{-2em}
    \caption{Additional Atari performance comparisons when using Pop-Art. See Figure~\ref{fig:atari-summary} for explanation. Note that in the 10-head scenario, Pop-Art is used to separately scale each head, leading to a clear benefit over the unscaled variant, but performance remains below return-based scaling.
    \vspace{-1em}
    }
    \label{fig:popart}
\end{figure*}

\begin{figure*}[t]
    \centering
    \includegraphics[width=\textwidth]{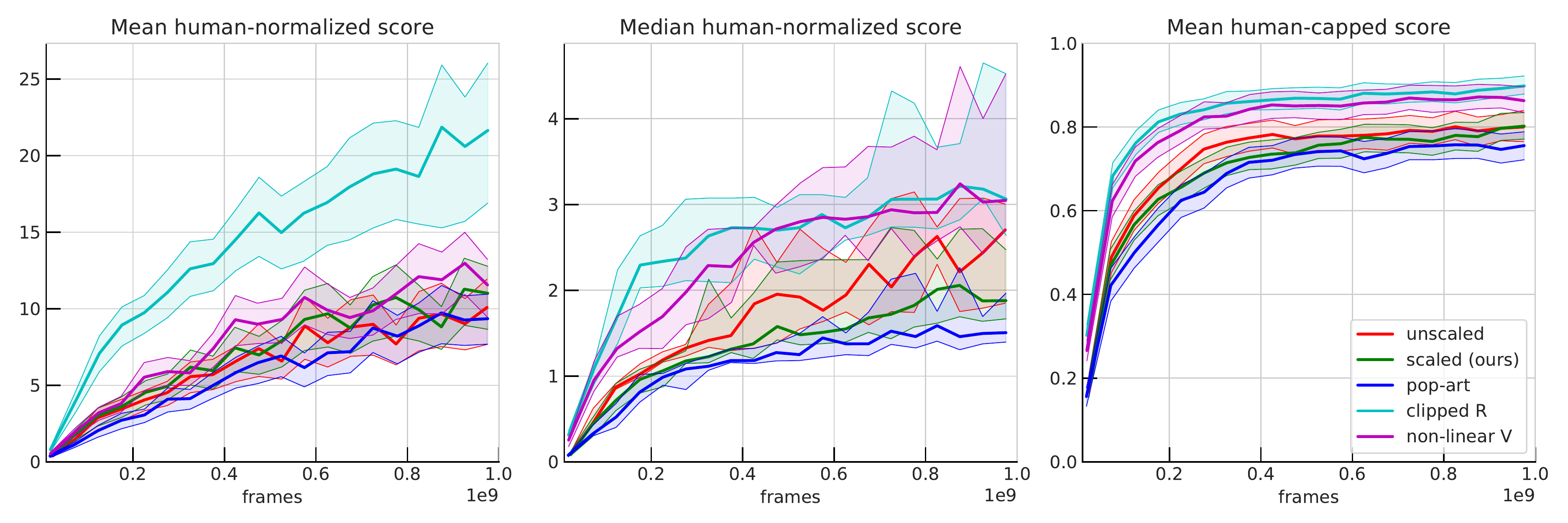}
    \vspace{-2em}
    \caption{Single-head Atari performance across various scaling methods. The best aggregate performance is attained by reward clipping (cyan), a method that does not preserve optimality (and is highly detrimental in some games) yet seems to be a surprisingly beneficial heuristic.
    \vspace{-1em}
    }
    \label{fig:1head}
\end{figure*}

\begin{figure*}[tb]
    \centerline{
    \includegraphics[width=\textwidth]{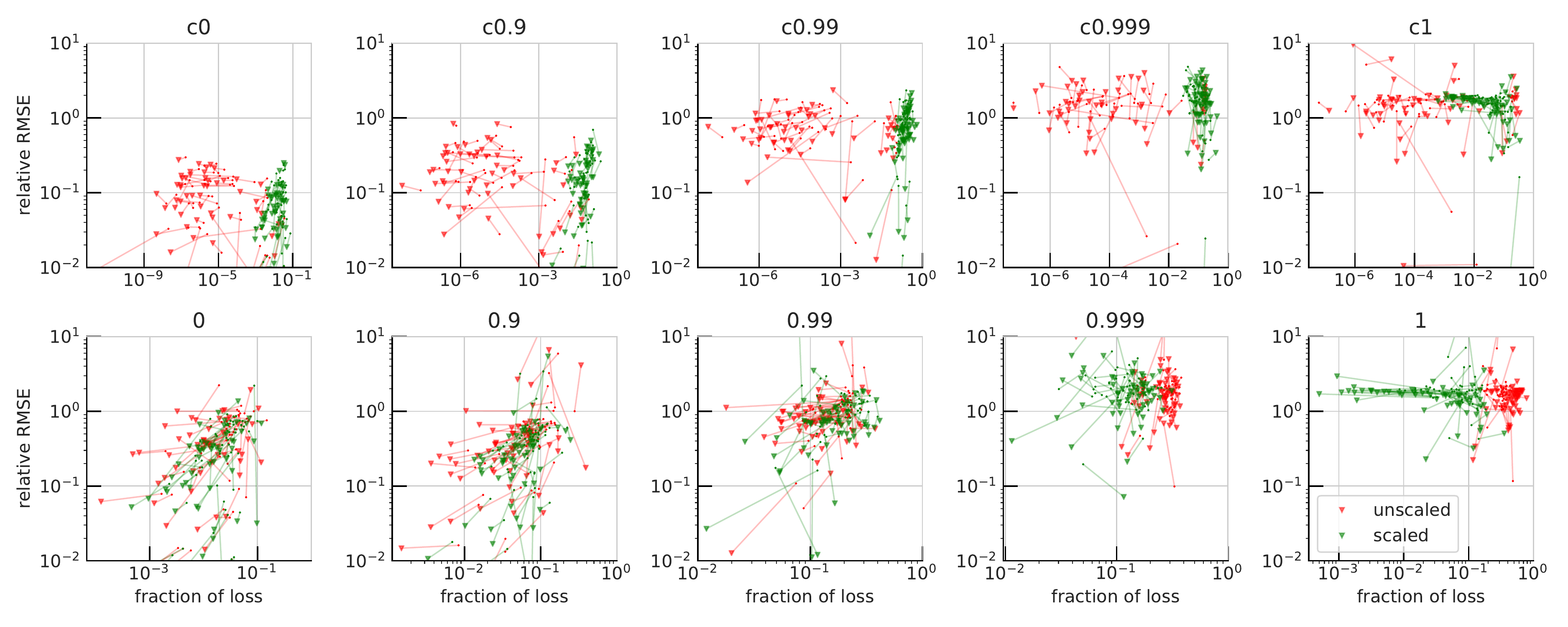}
    }
    \vspace{-1em}
    \caption{Fraction of the total loss that each of the 10 heads contributes, and how this relates to that heads value accuracy.
    Each of the 57 games is represented by a line segment with the circle and triangle showing the averages across the first 20\% and last 20\% of the run, respectively.
    Not surprisingly, the unscaled baseline (red) places most of the emphasis on unclipped and high-discount heads (bottom right), whereas for the scaled case, emphasis is balances across all heads. It is also worth highlighting that increased emphasis does not lead to improved accuracy, see also Figure~\ref{fig:imp-accuracy}.
    \vspace{-1em}
    }
    \label{fig:err-loss-frac}
\end{figure*}

\begin{figure*}[tb]
    \centering
    \includegraphics[width=0.95\textwidth]{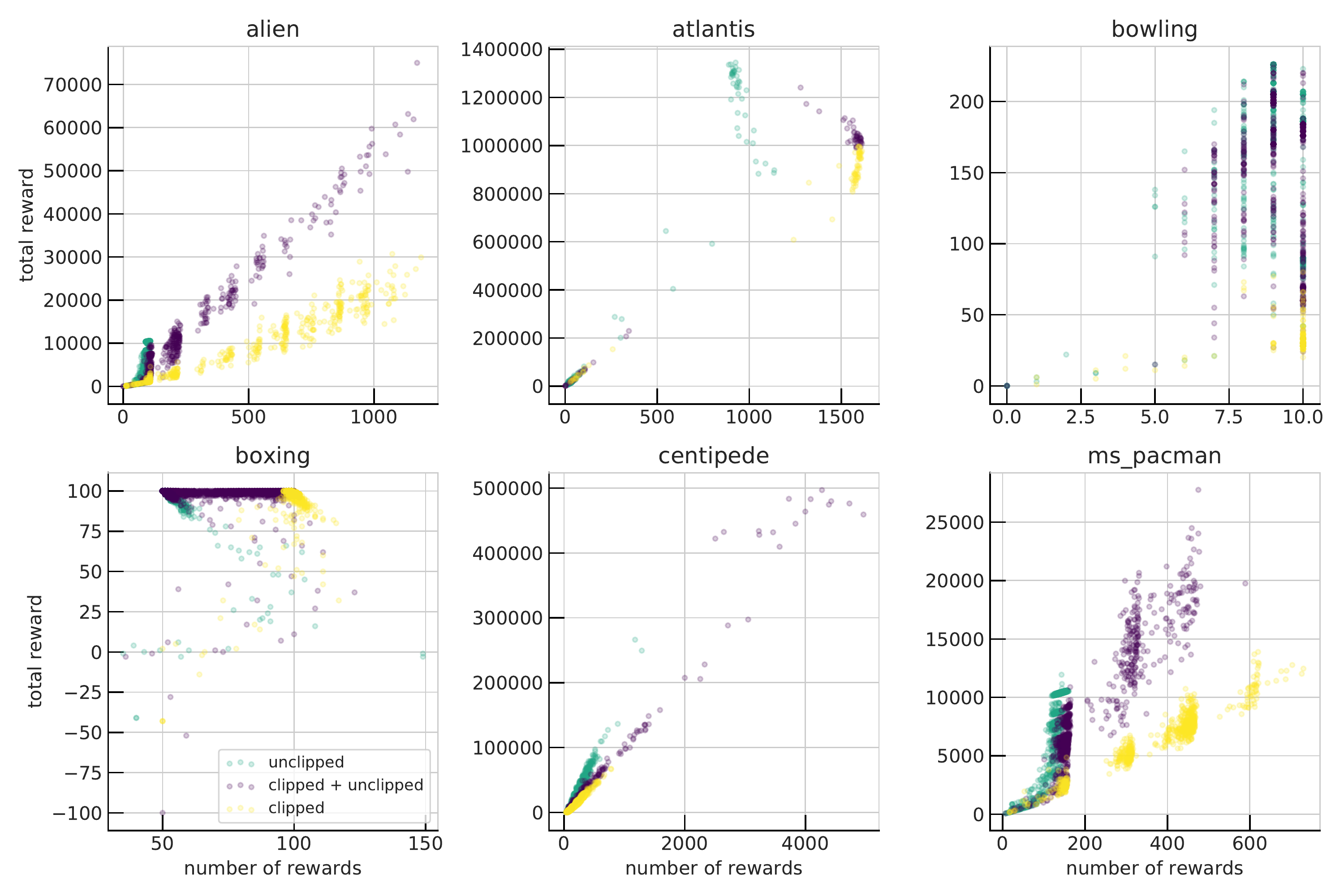}
    \vspace{-1em}
    \caption{Illustration of mismatching objectives when rewards are clipped (yellow) as compared to when they are unclipped (green), or the 10-head case where the agent can both pursue clipped and unclipped rewards (purple). Each point is a training episode, with total undiscounted reward (score) plotted as a function of number of non-zero rewards. One would expect a policy that pursues only clipped reward (yellow) to aim rightward, while a policy that maximises total score (green) would aim upward. These are some of the games where we do indeed observe such effects.
    However, as games like \textsc{Alien} or \textsc{Ms Pac-Man} show, the total score can nevertheless be higher for a policy that pursues clipped reward.
    \vspace{-1em}
    }
    \label{fig:clip-unclip}
\end{figure*}

\begin{figure*}[tb]
    \centering
    \includegraphics[width=\textwidth]{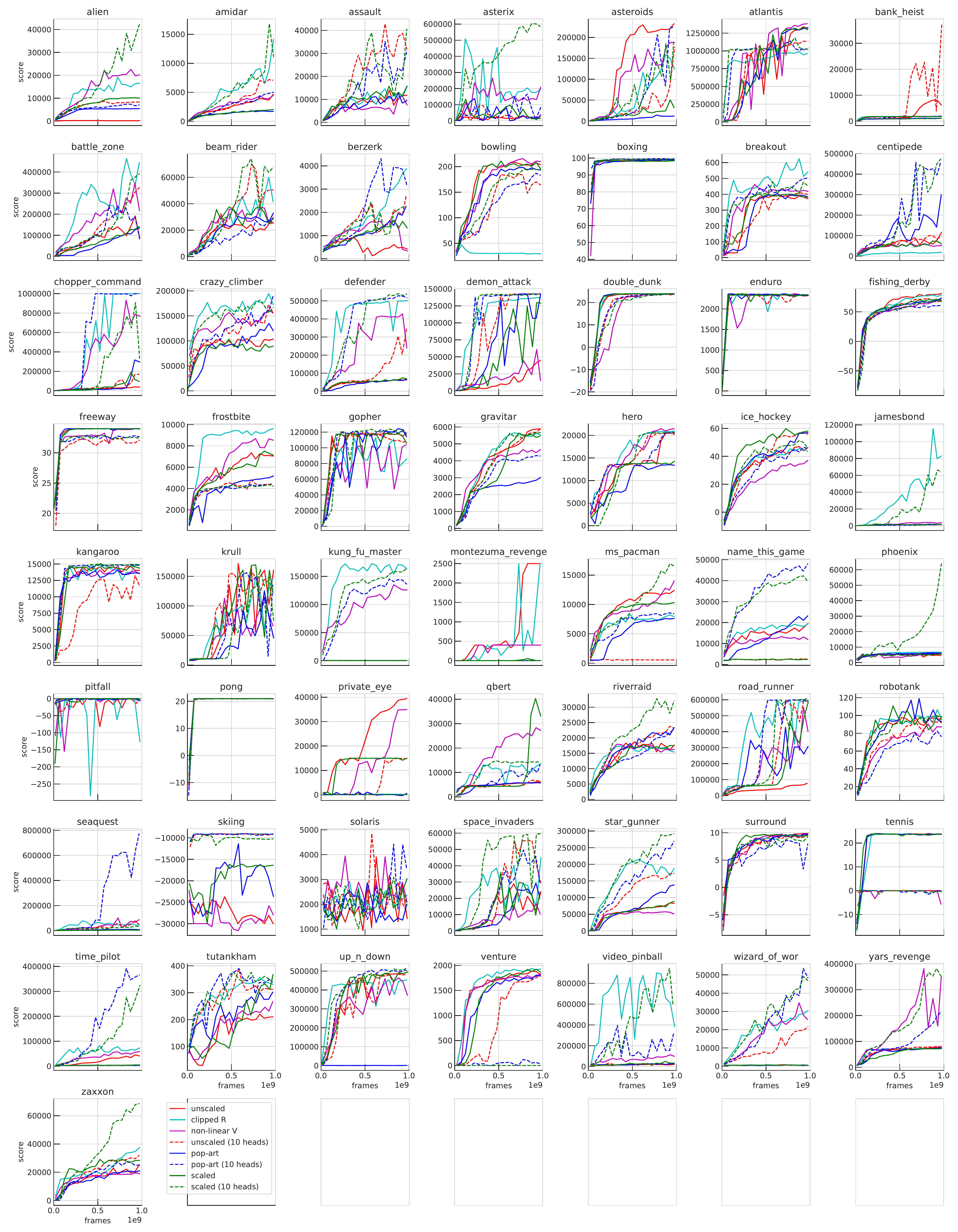}
    \vspace{-2em}
    \caption{Learning curves for all variants discussed in the paper (except the value bias initialisation from Figure~\ref{fig:skiing}), on all 57 Atari games.
    \vspace{-1em}
    }
    \label{fig:atari-all}
\end{figure*}

\section{Additional details}
In this section, we provide a couple of additional expressions that are tangential to the paper's main idea, but may be of interest to the reader.

\subsection{Edge cases}
\begin{itemize}[topsep=4pt,itemsep=0pt]
    \item For the regression setting of $T=1$ or $\gamma=0$, we obtain as expected $\sigma^2 = \V[R]$.
\item For the non-episodic setting ($T\rightarrow\infty$) we also obtain $\sigma^2 = \V[R]$, but only if 
$\E\left[G^2\right]$ does not grow with $T$.
\item In the case of constant (non-zero) rewards, $\V[R] = 0$, so 
\begin{equation*}
\sigma^2 = \V[\gamma] \E\left[G^2\right]\approx \frac{\V[\gamma]}{(1-\bar{\gamma})^2}\E[R]^2.
\label{eq:const-r}
\end{equation*}
\end{itemize}

\subsection{Episode boundaries}
For the (common) case where $\gamma_t$ is fixed to a constant $\gamma^\text{cst}$ throughout the episode (of average length $T$) and zero at the end, we have $\bar{\gamma} := \E[\gamma] =(1-\frac{1}{T}) \gamma^\text{cst}= \frac{T-1}{T} \gamma^\text{cst}$
and
\begin{eqnarray*}
\V[\gamma] &=& \left(\frac{T-1}{T} -\frac{(T-1)^2}{T^2}\right) \left(\gamma^\text{cst}\right)^2 
\\&=& \frac{T-1}{T^2} \left(\gamma^\text{cst}\right)^2
= \frac{\bar{\gamma}^2}{T-1}, 
\label{eq:var-gamma}
\end{eqnarray*}
which is approximately $1/T$ when $\gamma^\text{cst} \approx 1$.

\subsection{Brownian motion}
When there is no special structure in the returns (think shuffled rewards, themselves Gaussian-distributed), we can make the Brownian motion approximation and obtain
\[
\V[G] \approx \frac{1}{1-\bar{\gamma} } \V[R]
\]
where $\frac{1}{1-\bar{\gamma} }$ takes the role of a time horizon.

\end{document}